\pdfoutput=1

\documentclass[11pt]{article}

\usepackage[final]{acl}

\usepackage{times}
\usepackage{latexsym}
\usepackage{bm}
\usepackage[T1]{fontenc}

\usepackage[utf8]{inputenc}

\usepackage{microtype}

\usepackage{inconsolata}

\usepackage{graphicx}
\usepackage{multirow}
\usepackage{multicol}
\usepackage{booktabs}
\usepackage{amssymb}
\usepackage{amsmath}
\usepackage{xspace}
\usepackage{enumitem}
\usepackage{amssymb}
\usepackage{makecell}
\usepackage{amsmath}
\newcommand{\paratitle}[1]{\vspace{1.5ex}\noindent\textbf{#1}}
\newcommand{\ie}{\emph{i.e.,}\xspace}

\newcommand{\eg}{\emph{e.g.,}\xspace}

\newcommand{\ignore}[1]{}
\newcommand{\tabincell}[2]{\begin{tabular}{@{}#1@{}}#2\end{tabular}}
\title{LongReD: Mitigating Short-Text Degradation of Long-Context Large Language Models via Restoration Distillation}

\author{Zican Dong$^{1}$\thanks{Equal Contribution.},~Junyi Li$^{2*}$,~Jinhao Jiang$^{1}$,~Mingyu Xu$^3$,\\\textbf{Wayne Xin Zhao}$^1$\thanks{Corresponding author.},~\textbf{Bingning Wang}$^3$,
\textbf{Weipeng Chen}$^3$\\
        $^1$ Gaoling School of Artificial Intelligence, Renmin University of China \\ 
        $^2$ Department of Computer Science, National University of Singapore \\
        $^3$ Baichuan Inc.\\
        \texttt{dongzican@ruc.edu.cn, junyi\_cs@nus.edu.sg
        } \\\texttt{batmanfly@gmail.com, daniel@baichuan-inc.com}
}

\begin{document}
\maketitle
\begin{abstract}
Large language models (LLMs) have gained extended context windows through scaling positional encodings and lightweight continual pre-training. However, this often leads to degraded performance on short-text tasks, while the reasons for this degradation remain insufficiently explored. In this work, we identify two primary factors contributing to this issue: \emph{distribution drift} in hidden states and attention scores, and \emph{catastrophic forgetting} during continual pre-training. 
To address these challenges, we propose \textbf{Long} Context Pre-training with \textbf{Re}storation \textbf{D}istillation (\textbf{LongReD}), a novel approach designed to mitigate short-text performance degradation through minimizing the distribution discrepancy between the extended and original models. Besides training on long texts, LongReD distills the hidden state of selected layers from the original model on short texts. Additionally, LongReD also introduces a short-to-long distillation, aligning the output distribution on short texts with that on long texts by leveraging skipped positional indices. 
Experiments on common text benchmarks demonstrate that LongReD effectively preserves the model's short-text performance while maintaining comparable or even better capacity to handle long texts than baselines. Our code is available at \url{https://github.com/RUCAIBox/LongReD}.


\end{abstract}

\section{Introduction}
\label{sec:introduction}

\begin{figure}[htb]
    \centering
    \includegraphics[width=0.5\textwidth]{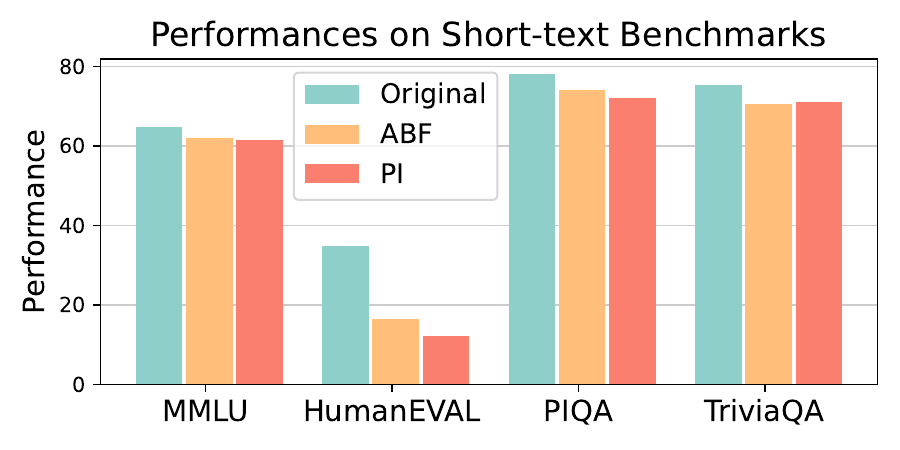}
    \caption{Comparison of the original and long-context models via ABF~\cite{xiong-naacl-2024-effective} or PI~\cite{chen-arxiv-2023-extending} on common short-text benchmarks.}
    \label{fig:drop}
\end{figure}
Large language models (LLMs) have exhibited remarkable performance across a wide range of text and multimodal tasks~\cite{brown-nips-2020-gpt3,openai-arxiv-2023-gpt4,zhao-arxiv-2023-survey,Touvron-arxiv-2023-llama2,Dubey-arxiv-2023-llama3,tang-naacl-2025-dawnicl,wang-2025-arxiv-unvieling,chen-arxiv-2025-empirical}. However, their abilities to process long contexts are constrained by the positional encodings and the attention mechanism, which define the context window size based on the length of the pre-training data~\cite{su-neurocomputing-2024-roformer,press-iclr-2022-alibi}. When the input text exceeds this context window, the model encounters out-of-distribution positional information, resulting in significant performance degradation~\cite{chen-arxiv-2023-extending,peng-arxiv-2023-yarn,dong-arxiv-2024-exploring}.


To address the challenges of long context processing, various methods have been proposed to extend the context window of LLMs~\cite{bloc97-reddit-2023-ntk,xiao-arxiv-2023-streaming,xiong-naacl-2024-effective,chen-arxiv-2023-extending}. A prominent approach involves scaling positional encodings combined with lightweight continual pre-training. Leveraging the properties of RoPE~\cite{su-neurocomputing-2024-roformer}, these methods interpolate positional indices or adjust the RoPE base to prevent out-of-distribution rotary angles beyond the original context window~\cite{chen-arxiv-2023-extending, ding-icml-2024-longrope, peng-arxiv-2023-yarn,xiong-naacl-2024-effective}. Subsequently, lightweight continual pre-training adapts LLMs to the extended context and modified encodings, enabling context window up to 128K or even 1M tokens~\cite{xiong-naacl-2024-effective,fu-icml-2024-data,Zeng-arxiv-2024-glm}. 
However, as shown in Figure~\ref{fig:drop}, these long context window extension techniques come at the cost of degraded performance on short-text tasks~\cite{xiong-naacl-2024-effective,Dong-COLING-2024-BAMBOO}. Yet, the root causes of this performance decline and potential mitigation strategies remain under-explored.

In this study, we aim to demystify the factors contributing to the performance degradation of short-text tasks after context window extension. Our analysis identifies two critical factors: \textbf{distribution drift} and \textbf{catastrophic forgetting}. The key findings of our work include: (1) Continual pre-training seeks to recover the original model's internal distribution, but the restoration is inherently imperfect. (2) Distribution shift in hidden states potentially leads to performance degradation. (3) During continual pre-training, the performance on short-text tasks initially improves but subsequently declines as training progresses, highlighting the presence of catastrophic forgetting. (4) Replaying short text data
is effective in mitigating forgetting and improving performance stability.


Based on our observations, we propose a novel approach called \textbf{Long} Context Pre-training with \textbf{Re}storation \textbf{D}istillation (\textbf{LongReD}) to mitigate the degradation in short-text capabilities of long-context LLMs. The central idea is that the short-text capacities of the extended model can be better preserved if it accurately simulates the original distributions before extension. To achieve this, in addition to typical \textbf{long-text training}, we introduce a \textbf{short-text distillation} objective, which employs the original model as a teacher to distill its hidden states on short texts into the extended model. This training objective minimizes distribution drift and alleviates catastrophic forgetting.
Moreover, we propose a \textbf{short-to-long distillation} training objective to bridge the gap between short-text distillation and long-text training. In this setup, the original and extended models are fed with normal positional indices and skipped positional indices, respectively. 
By applying the distillation on the output distributions of the last layer, the short-text capacities can be effectively transferred and integrated into long-text processing.

To the best of our knowledge, this work represents the first systematic analysis of the reasons behind the performance degradation of long-context LLMs on short-text tasks. Furthermore, we propose a general method to mitigate this short-text degradation. To assess the effectiveness of our method, we extend the context window of Llama-3-8B and Mistral-7B-v0.3 and assess their performance on both short-text and long-text tasks. Experimental results demonstrate that our method preserves the original models' performance on short-text tasks while maintaining or even improving their long-context modeling capabilities.



\section{Background}
\label{sec:background}

\subsection{Transformer}
Owing to strong capabilities and scalability, Transformer decoders have served as the backbone of most LLMs~\cite{vaswani-nips-2017-attention, Touvron-arxiv-2023-llama2,Dubey-arxiv-2023-llama3,tang-aaai-2025-unleashing}. Given a Transformer decoder with $L$ layers and $N$ heads and an input sequence $\bm{x} = \{ x_1, \dots, x_T \}$ consisting with $T$ tokens, the output of the $l$-th layer can be denoted as $\mathbf{H}_l = \{\bm{h}_{l,1}, \dots, \bm{h}_{l,T}\}$. Each Transformer layer consists of a multi-head attention (MHA) module and a feed-forward network (FFN) module with residual connections connecting them, as shown by the following formula:
\begin{align}
    \widetilde {\mathbf H}_l &= \mathrm{MHA}(\mathbf H_{l-1})+\mathbf H_{l-1},\\\mathbf H_l &= \mathrm{FFN}(\widetilde {\mathbf H}_l) + \widetilde {\mathbf H}_l.
\end{align}

In the $i$-th head of the MHA module at the $l$-th layer, the hidden states are first projected into query, key, and value matrices, \ie $\mathbf{Q}_l^{i}$, $\mathbf{K}_l^{i}$, and $\mathbf{V}_l^{i}$. Positional information is then incorporated into the query and key matrices through the RoPE with the rotation matrix $\mathbf{R}_{\theta}$ ($\theta$ is the RoPE base)~\cite{su-neurocomputing-2024-roformer}. These matrices are subsequently processed through a dot product followed by a softmax operation to compute the attention scores $\mathbf{A}_l^{i}$. Finally, the value representations are weighted by the attention scores, and all attention heads are concatenated and projected to produce the attention output of the $l$-th layer as follows:
\begin{align}
    \mathbf A_l^{i} &= \mathrm{Softmax}({\mathbf Q_{l}^{i}}^\intercal \mathbf R_{\theta} {\mathbf K_{l}^{i}}/\sqrt d ),\\
    \mathrm{MHA}(\mathbf H_{l-1}) &= \mathrm{Concat}(\{\mathbf A_l^{i} \mathbf V_l^{i}\}_{i=1}^N)\mathbf W^O,
\end{align}
where $d$ is the dimension of query and key, $\mathbf W^O$ is the projection matrix, and $\mathrm{Concat}$ is the concatenation of hidden states.

\subsection{Measures of Distributional Discrepancy}
\label{sec:metric}

After extending the context window, the parameters of LLMs undergo changes, resulting in a shift in the distribution of hidden states. To quantify such distributional discrepancy, we propose two measures, \ie \textbf{hidden state similarity} and \textbf{attention Kullback-Leibler (KL) divergence}. Specifically, given the hidden states $\mathbf{H}_l$, $\mathbf{\widehat{H}}_l$ from the original and extended models, hidden state similarity measures the average cosine similarity between hidden states at the $l$-th layer across all positions. Similarly, attention KL divergence calculates the average KL divergences between the $i$-th attention head distributions $\mathbf{A}^i_l$, $\mathbf{\widehat{A}}^i_l$ from the original and extended models at the $l$-th layer. Higher hidden state similarity or lower attention KL divergence indicates less discrepancy between the two models. These metrics are formulated as follows:
\begin{align}
    \operatorname{Sim}(\mathbf{H}_l, \mathbf{\widehat{H}}_l) &= \frac 1 T\sum_{t=1}^T \frac { {\bm{h}_{l,t}}^\intercal \bm{\widehat{h}}_{l,t}}{\lVert \bm h_{l,t}\rVert\lVert \bm{\widehat{h}}_{l,t}\rVert},\\
    \operatorname{KL}(\mathbf{A}_l^{i},\widehat{\mathbf{A}}_l^{i}) &= \frac 1 T \sum_{t=1}^T \sum_{j=1}^t a^{i}_{l,t,j} \log \frac {a^{i}_{l,t,j}}{\hat a^{i}_{l,t,j}}.
\end{align}



        

\section{Empirical Analysis}
\label{sec:analysis}

Prior work~\cite{xiong-naacl-2024-effective,ding-icml-2024-longrope} demonstrated that the short-text capacities will degrade after lightweight continual pre-training for extending the context window of LLMs. However, the secrets behind the performance decline have not been fully explored. In this section, through conducting empirical analysis, we attribute the performance degradation to two factors, \ie \emph{distribution drift} and \emph{catastrophic forgetting}.



\subsection{Distribution Drift}
\label{sec:drift}




\paratitle{Imperfect Distribution Restoration.}
In previous work~\citep{chen-arxiv-2023-extending, ding-icml-2024-longrope}, RoPE configurations are simply modified to extend the context window, undoubtedly leading to the change of distributions. Although continual pre-training is proposed to adapt LLMs to the extended context, a natural question is whether continual pre-training can eliminate the distributional discrepancy. To verify this, we modify 
RoPE base to $2e7$ and $1e8$, and train Llama-3-8B (with context window size of 8192 tokens) on 1B tokens with the sequence length of 32K and 128K, respectively, denoted as Llama-3-8B-32K and Llama-3-8B-128K.  Besides, we select Llama-3-8B-Instruct-262K~\cite{gradient-hf-2024-longcontextllama3}, which is trained on Llama-3-8B-Instruct. 
To measure the distributional discrepancy, we employ these models before and after continual pre-training for inference on 1000 test samples with the length of 8192 tokens from SlimPajama~\cite{soboleva-huggingface-2023-slimpajama}. The results of hidden state similarity and attention KL divergence are shown in Table~\ref{tab:restoration}. As can be seen, as the 
RoPE base increases, the extended LLMs before continual pre-training exhibit lower hidden state similarity and higher KL divergence from the original models. After training on long-context data, the similarity increases, and the KL divergence declines to some extent. However, \textbf{despite the LLM striving to restore the inner distribution resembling the original model during continual pre-training, there still exists a certain degree of distributional discrepancy.}


\begin{table}[htb]
    \centering
    \resizebox{\linewidth}{!}{
    \begin{tabular}{lccccc}
    \toprule
         \multirow{2}{*}{Model} & \multirow{2}{*}{Base} & \multicolumn{2}{c}{Simi $\uparrow$} & \multicolumn{2}{c}{KLD($\times 10^{-5}$) $\downarrow$}   \\
         & & before &after&before&after \\\midrule
         Llama-3-8B-32K&2e7& 0.92&0.95&3.51&1.51\\
         Llama-3-8B-128K&1e8&  0.83&0.94&6.74&1.68\\
         Llama-3-8B-Ins-262K&2.8e8& 0.76&0.91&8.75&2.50\\\bottomrule
    \end{tabular}}
        \caption{Results of hidden state similarity and KL divergence before and after continual pre-training.}
    \label{tab:restoration}
\end{table}

\begin{figure}[htb]
   \centering
    \includegraphics[width=\linewidth]{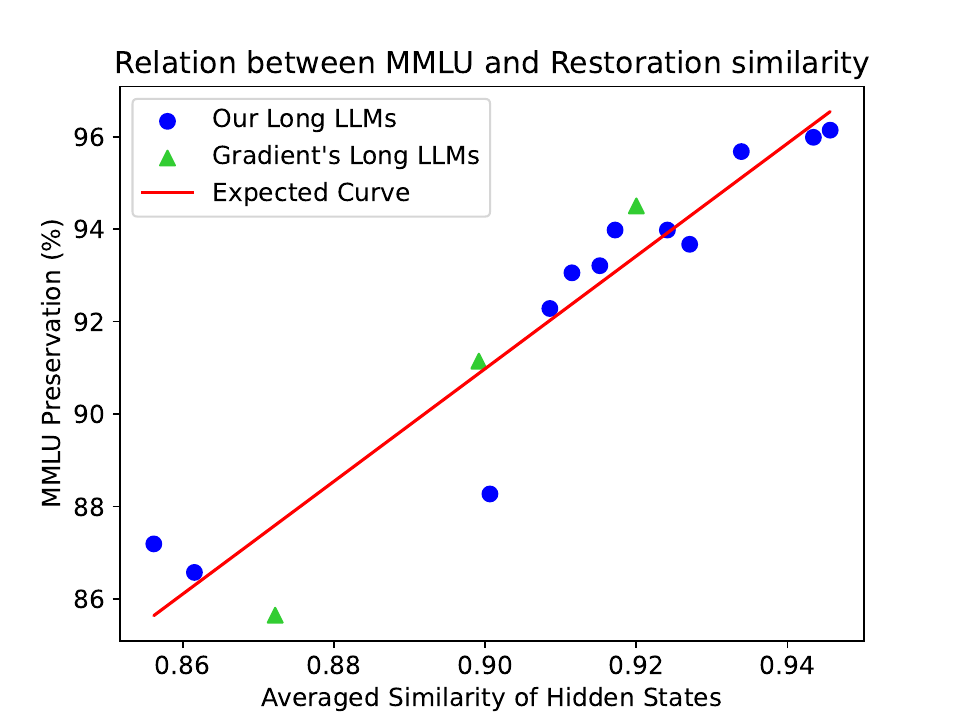}
    \caption{Relationship between MMLU performance preservation of long-context models \emph{w.r.t.} the hidden states similarity.}
    \label{fig:mmlu_restoration}
    \vspace{-0.2cm}
\end{figure}

\paratitle{Relationship between Distribution Drift and Performance Degradation.} Owing to the distribution drift of the extended model, the inner working mechanisms may also change when processing the original short text data. 
To answer the question of whether the distributional discrepancy contributes to the performance decline in short text, we train the Llama-3-8B model using various context window extension methods and training strategies (detailed in Appendix~\ref{app:drift_models}). Additionally, we incorporate existing open-source long-context models, which were extended from Llama-3-8B-Instruct by Gradient AI~\cite{gradient-hf-2024-longcontextllama3}. We compute the average hidden state similarity across all layers between these extended models and their original counterparts and assess their overall performance on the MMLU benchmark (mainly focused on short-text tasks)~\cite{Hendrycks-iclr-2021-mmlu}. Then, we compute the MMLU performance preservation ratio of long-context models compared to the original models \emph{w.r.t.} the hidden state similarity between them, as shown in Figure~\ref{fig:mmlu_restoration}. Generally, extended models that maintain higher similarity to their original versions preserve more short-text performance. This finding highlights that \textbf{distribution shifts in extended models are a significant factor contributing to performance degradation in short text.}

\subsection{Catastrophic Forgetting}
\label{sec:forgetting}

Catastrophic forgetting is always a critical problem during continual pre-training~\cite{wu-arxiv-2024-continual}. Owing to the distributional difference of short texts and long texts, adapting for long context window may unavoidably leads to a trade-off on short texts. 
In this section, we examine the effect of the training steps and training data on the short-text performances to verify the forgetting phenomenons. 

\begin{figure}[htb]
    \centering
    \includegraphics[width=
\linewidth]{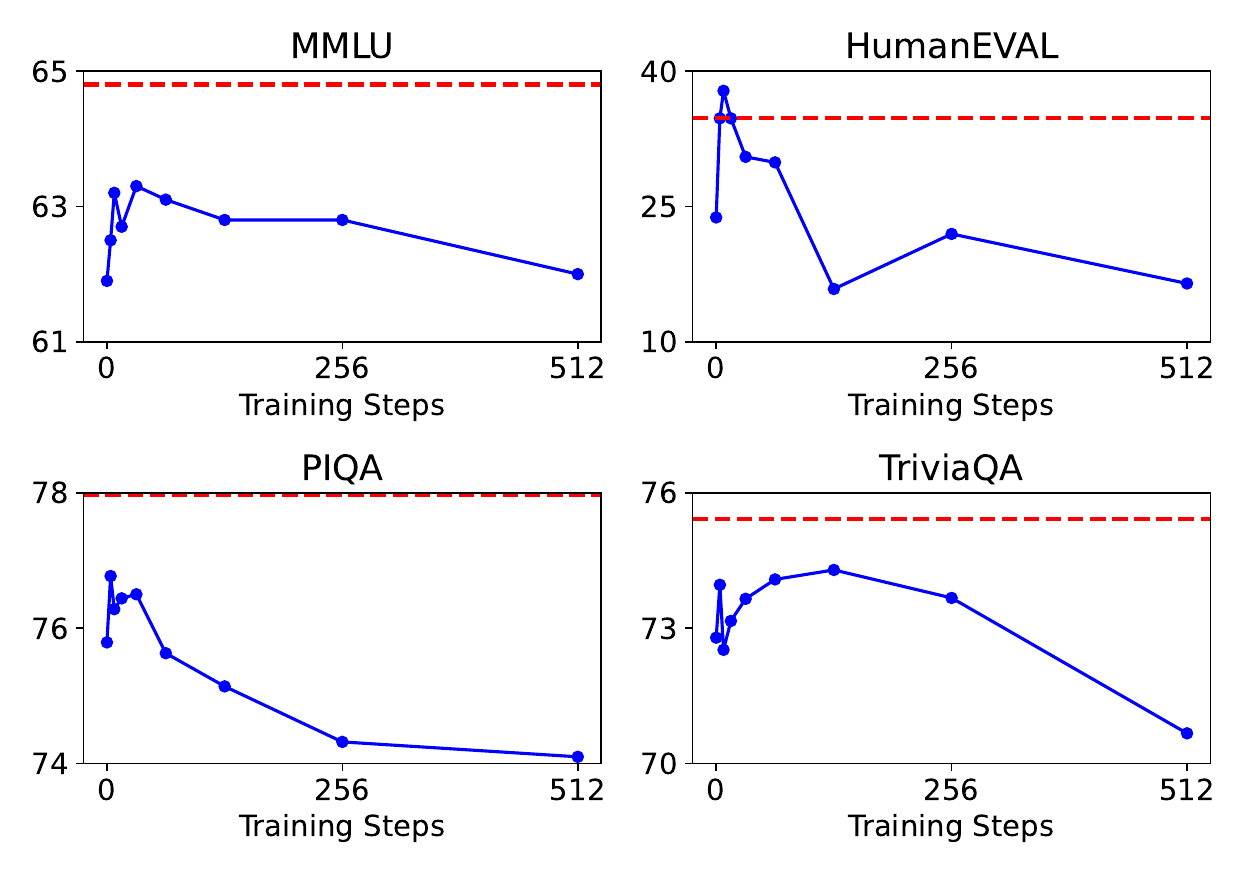}
    \caption{Results of models with different training steps.}
    \label{fig:step}
\end{figure}

\paratitle{Effect of Training Steps.} We train Llama-3-8B with the RoPE base of $2e7$ on the SlimPajama dataset and obtain checkpoints with different training steps, where the batch size is $64$ and the training length is 32K. Then, we evaluate the models on four short-text benchmarks, \ie MMLU~\cite{Hendrycks-iclr-2021-mmlu}, HumanEval~\cite{chen-arxiv-2021-humaneval}, TriviaQA~\cite{Joshi-acl-2017-triviaqa}, and PIQA~
\cite{bisk-aaai-2020-piqa}. Figure~\ref{fig:step} presents the performance change with different training steps. It is clear that the performance is restored at the initial steps (usually less than 32 steps) in spite of fluctuations in some datasets. However, the performance gradually drops with the increase of training steps, which demonstrates that\textbf{ the continual adaptation for long context window will lead to catastrophic forgetting issues on short-text tasks.}

\paratitle{Effect of Training Data Mixture.} Beyond the training steps, we also evaluate the effect of mixtures of training data. 
We replace half of the long text data with short text data with a length of 8K. We train Llama-3-8B on 1B tokens and evaluate them on different benchmarks. As shown in Table~\ref{tab:data_effect}, the length of training data plays a critical role in downstream tasks. 
Introducing short-text data achieves better short-text performances than only training on long-text data. This verifies that \textbf{long texts are critical for forgetting problems, which can be mitigated by replaying short texts.}



\begin{table}[htb]
    \centering
    \resizebox{\linewidth}{!}{
    \begin{tabular}{l|cccc}
    \toprule
     Length & MMLU & HUMANEVAL & PIQA &TriviaQA\\\midrule
       Long & 62.0&14.02& 74.10 & 70.67\\
        Long+Short & 62.5&16.46& 78.24 & 72.82 
         \\\bottomrule
    \end{tabular}}
    \caption{Results of models trained with data mixtures with different lengths.}
    \label{tab:data_effect}

\end{table}

%
\section{Approach}
\label{sec:method}

\begin{figure*}[htb]
    \centering
    \includegraphics[width=\linewidth]{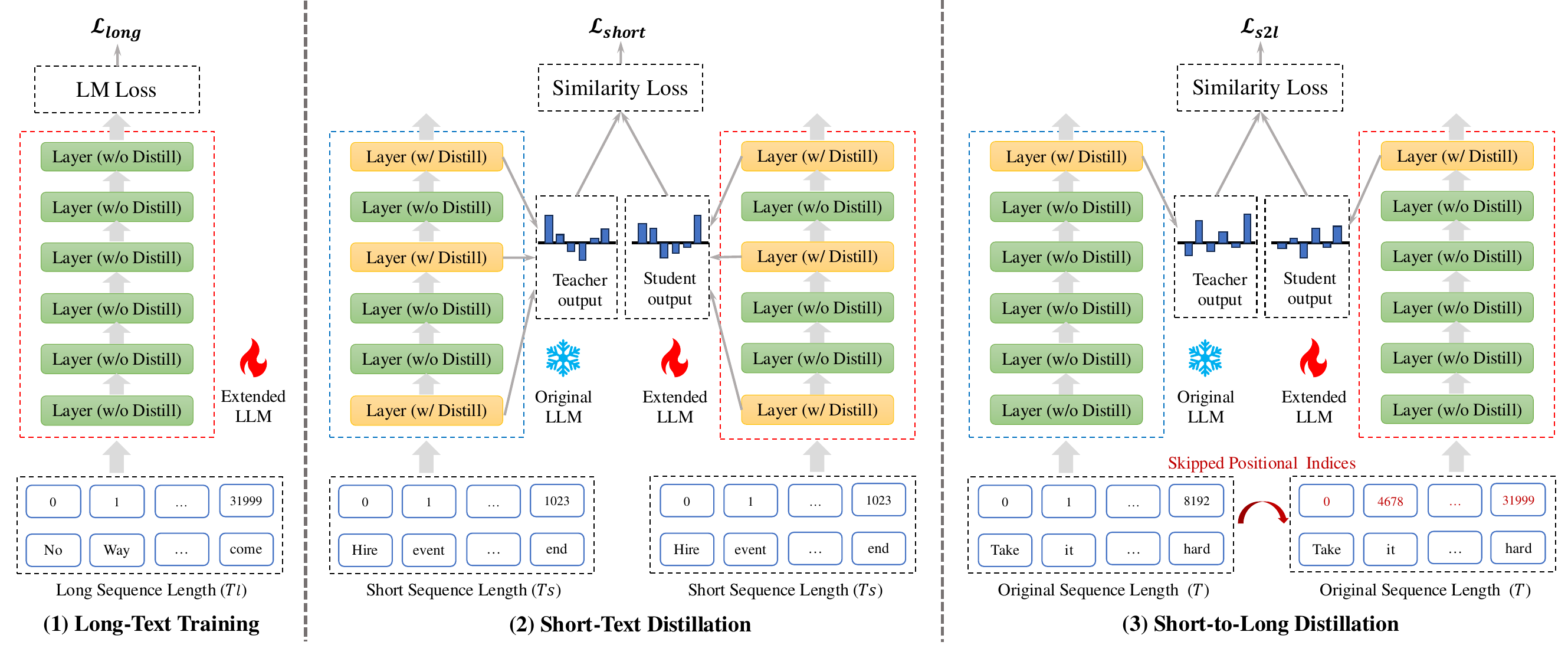}
    \caption{Overview of our proposed Long Context Pre-training with Restoration Distillation (LongReD). The method consists of three parts, \ie long-text training, short-text distillation, and short-to-long distillation.}
    \label{fig:architecture}
    \vspace{-0.3cm}
\end{figure*}

Inspired by the above findings, we propose \textbf{Long} Context Pre-training with \textbf{Re}storation \textbf{D}istillation (\textbf{LongReD}), a novel context window extension framework to improve short-text capacities of long-context LLMs through decreasing distribution drift and mitigating catastrophic forgetting. Unlike only training on long texts, our strategy combines three different training objectives at each training step, \ie \textbf{long-text training}, \textbf{short-text distillation}, and \textbf{short-to-long distillation}. For the three objectives, we employ three datasets $\mathcal D_1,\mathcal D_2,\mathcal D_3$ with different lengths $T_l$, $T_s(<T)$, and $T$, respectively, where $T$ and $T_l$ are the original and target long context window length, and $T_s$ is the short text length. 
The overall architecture is displayed in Figure~\ref{fig:architecture}.

\subsection{Long Text Training}

To extend the context window of an LLM, we follow previous work to first scale positional encodings via ABF~\cite{xiong-naacl-2024-effective} or PI~\cite{chen-arxiv-2023-extending} and directly continually training on the long text dataset. Given any long text $\bm{x}$ from the dataset $\mathcal D_1$, the model $\Theta_e$  is trained with the language modeling objective to minimize the cross-entropy loss $\mathcal{L}_{long}$ as follows:
\begin{equation}
    \mathcal{L}_{long} = - \sum_{\bm{x}\in \mathcal D_1} \frac 1 {T_{l}}\sum_{t=1}^{T_{l}} \log \operatorname{Pr}(x_t |\bm{x}_{<t}; \Theta_{e}).
\end{equation}
Through training on long texts, the model can learn to model long-term dependencies and adapt to the extended context window.

\subsection{Short-Text Distillation}
Section~\ref{sec:analysis} demonstrates the negative effect of distribution drift and catastrophic forgetting on short-text capacities while only training on long texts. Though replaying short text data can partly mitigate the issue, the performance restoration is limited. Therefore, we propose short-text distillation, a knowledge distillation method to 
minimize the distribution discrepancy between the extended model $\Theta_e$ and the original model $\Theta_o$. 

\paratitle{Distillation Layer Selection.} To decrease the distribution discrepancy, we choose to distill the inner hidden states of LLMs. Yet, ensuring all the hidden states are the same as the original model is impossible and may significantly decrease the long text modeling capacities. Thus, we only select  $M(<L)$ layers as the distill layers $\mathbf{L}_M=\{l_1, \dots, l_M\},\forall l_i\in \{0,\dots, L\}$ and distill the outputs of these layers. Specifically, we propose an attention-based layer selection method where we first compute the KL divergence of attention scores between the extended and original models and then select layers with the largest KL divergences for distillation~\cite{men-arxiv-2024-shortgpt,dong-arxiv-2025-domain}. We also select the critical last layer that directly determines the output.



\paratitle{Short-Text Distillation Loss.}
Given an input sample $\bm{x}$ from the short text dataset $\mathcal D_2$, we input the sample into the two models with its original positional indices $\bm p=[0,\dots, T_s-1]$. Then, we obtain the output hidden states of the selected layers of the two models, \ie $\{\mathbf H_{l_1; \bm p, \Theta_{e}},\dots, \mathbf H_{l_M; \bm p, \Theta_{e}}\}$ and $\{\mathbf H_{l_1; \bm p, \Theta_{o}},\dots, \mathbf H_{l_M; \bm p, \Theta_{o}}\}$. We 
compute the cosine similarity between the hidden states of the two models at the same layer and position and sum up them as the short-text distillation loss $\mathcal{L}_{short}$,  formulated as follows:
\begin{equation}
    \mathcal{L}_{short} = -\sum_{\bm{x}\in \mathcal D_2} \sum_{i=1}^M\operatorname{Sim}( \mathbf H_{l_i; \bm p, \Theta_{e}}, \mathbf H_{l_i; \bm p,\Theta_{o}}).
\end{equation}

\subsection{Short-to-Long Distillation}
\label{sec:s2l_distillation}


The two objectives of long-text training and short-text distillation separately optimize the model on long and short texts, resulting in a challenging performance trade-off between long and short texts. We argue that long and short texts share some inherent linguistic commonalities. Thus, we propose an additional short-to-long distillation to transfer the short-text capabilities to long texts, further bridging the gap between short and long texts.


\paratitle{Skipped Positional Indices.} To distill short-text capabilities to long texts, we take the short text with the original context window size $T$ as input but employ the skipped positional indices method, which can simulate long-text positions and capture longer-term dependencies with short texts~\cite{zhu-iclr-2024-pose,wu-nips-2024-cream}.
To be specific, for a sequence $\bm{x}$ drawn from $\mathcal{D}_3$, we first define a threshold $T_b$ to split the positional indices into three parts, \ie $\bm p_{head}$, $\bm p_{mid}$, and $\bm p_{tail}$:
\begin{align}
    \bm p  &= \bm p_{head}\cup\bm p_{mid}\cup\bm p_{tail},\\ \bm p_{head} &= \{0, \dots, T_b-1\},\\ \bm{p}_{mid} &= \{T_b, \dots, T-T_b-1\},\\\bm p_{tail} &= \{T-T_b, \dots, T-1\}.
\end{align}
Then, we modify the positional indices of the three segments. We keep $\bm p_{head}$ unchanged but modify the end positional index of $\bm p_{tail}$ to be equal to the target long text length $T_l$. For the middle segment, we employ uniform sampling or CREAM~\cite{wu-nips-2024-cream} which upsamples middle positions to determine the end positional index of $\bm p_{mid}$ as $T_{me}\in \{T-T_b,\dots,  T_l-T_b-1\}$. And the start positional index of $\bm p_{mid}$ is equal to $T_{me}$ minus the length of this middle segment $T - 2T_b$, \ie $T_{me} - T + 2T_b$. The modified positional indices of the input text are formulated as follows:
\begin{align}
    \hat{\bm{p}}  &= \hat{\bm{p}}_{head}\cup\hat{\bm{p}}_{mid}\cup\hat{\bm{p}}_{tail},\\ 
    \hat{\bm{p}}_{head} &= \{0, \dots, T_b-1\},\\
    \hat{\bm{p}}_{mid} &=\{T_{me}-T+2T_b,\dots, T_{me}\},\\
    \hat{\bm{p}}_{tail} &= \{T_l-T_b, \dots, T_l-1\}.
\end{align}
Details of the skipped positional indices methods are introduced in Appendix~\ref{app:cream}.

\paratitle{Short-to-Long Distillation Loss.}
After obtaining skipped positional indices $\hat{\bm {p}}$, we input these indices into the extended model $\Theta_e$ while using the normal positional indices $\bm p$ for teacher model $\Theta_o$. Then, we obtain the outputs of the $L$-th last layer of both models and compute the cosine similarity between them. Then, we regard the negative cosine similarity as the loss to distill the output distribution into longer positions. 
The training objective $\mathcal{L}_{s2l}$ can be represented as follows:
\begin{equation}
    \mathcal{L}_{s2l} = -\sum_{\bm{x}\in \mathcal D_3}  \operatorname{Sim}( \mathbf H_{L; \hat{\bm{p}}, \Theta_{e}}, \mathbf H_{L; \bm p,\Theta_{o}}).
\end{equation}
Notably, we only distill the output of the last layer instead of selected layers in short-text distillation to avoid the disturbance of latent positional information in the middle layers. 

\subsection{Joint Training Objective}

Finally,  we aggregate losses from these three training objectives. For each batch of training data, we sample data points from the three datasets with a fixed ratio. Then, we independently compute loss through the three different objectives on the corresponding data and sum them up to balance the long and short text capacities. The final loss $\mathcal{L}_{final}$ can be represented as follows:
\begin{equation}
    \mathcal{L}_{final} = \mathcal{L}_{long} + \alpha_1 \mathcal{L}_{short} + \alpha_2 \mathcal{L}_{s2l},
\end{equation}
where $\alpha_1$ and $\alpha_2$ are hyper-parameters that control the proportion of the three parts.
\section{Experiment}
\label{sec:experiment}


\begin{table*}
    \centering
    \resizebox{\textwidth}{!}{
    \begin{tabular}{l|cccccccccc|c|c} \toprule
         Model&  CW&  PE&  Data&  Method&  General&  Coding&  Math&  RC&   Common&Short Avg.&RULER &Avg.\\\midrule
 \multirow{13}{*}{\makecell{Llama-3\\-8B}}& 8K& -& -& -& 68.18& 41.20& 30.12& 63.78& 72.54& 55.16&- &-\\ \cmidrule{2-13}
 & 32K& ABF& Long& CPT& 65.39& 29.32& 29.50& 60.98& 69.83& 51.00&82.80&56.30\\ 
 & 32K& ABF& Mix& CPT&66.04 & 27.40& 29.32& 34.65& 71.88& 45.86&\textbf{85.04} &52.39\\
 & 32K& ABF& Mix& LongReD-C& \textbf{67.65}& \textbf{39.90}& \textbf{31.52}& \textbf{62.84}& 72.35& \textbf{54.85}&84.98&\textbf{59.87}\\
 & 32K& ABF& Mix& LongReD-U& 67.05& 39.71& 30.90& 62.51& \textbf{72.61}&  54.56 &84.08&59.48\\\cmidrule{2-13}
 & 32K& PI& Long& CPT& 63.84& 27.13& 26.18& 61.95& 67.71& 49.36&\textbf{82.17} &54.83\\ 
 & 32K& PI& Mix& CPT& 64.22& 27.29& 29.16& 36.63& 71.09& 45.68&81.63&51.67\\ 
 & 32K& PI& Mix& LongReD-C& \textbf{66.45}& {38.39}& {30.78}& \textbf{62.79}& {71.30}& 53.94&81.19&58.48\\
 & 32K& PI& Mix& LongReD-U& 66.34&\textbf{ 38.83}& \textbf{31.68}& 62.76& \textbf{72.36}& \textbf{54.39}&79.30&\textbf{58.54}\\\cmidrule{2-13}
 & 128K& ABF& Long& CPT& 64.06 & 30.04& 27.72& 60.49& 68.37& 50.14&\textbf{69.70} &53.40\\
 & 128K& ABF& Mix& CPT& \textbf{66.39}& 14.03& 29.02& 35.28& \textbf{71.91}& 43.33&69.64&47.71\\
 & 128K& ABF& Mix& LongReD-C& 65.99& \textbf{40.03}& \textbf{30.26}& 62.39& 71.50& \textbf{54.03}&64.93&55.85\\
 & 128K& ABF& Mix& LongReD-U& 65.61& 39.87& 29.67& \textbf{62.47}& 71.64& 53.85&68.41&\textbf{56.28}\\
 \midrule
 \multirow{5}{*}{\makecell{Mistral\\-7B-v0.3}}& 32K& -& -& -& 65.36& 33.20& 26.98
& 68.07& 61.86& 51.09& -&-\\\cmidrule{2-13}
  & 128K& ABF& Long& CPT& 55.29& 26.37& 15.79
& 52.93& 53.07& 40.68& 44.63&41.35\\
 & 128K& ABF& Mix& CPT& 55.99& 26.36& 15.30& 60.21& 52.74& 40.66& 45.76&41.51\\
 & 128K& ABF& Mix& LongReD-C& 61.45& 26.66& \textbf{19.88}
& 65.25& 58.26& 46.30& \textbf{58.37}&48.31\\
  & 128K& ABF& Mix& LongReD-U& \textbf{62.42}& \textbf{29.64}& 19.56& \textbf{66.86}& \textbf{59.98}& \textbf{47.69}& 53.60&\textbf{48.68}\\
 \bottomrule
    \end{tabular}}
    \caption{Comparison of performances of short-text and long-text benchmarks of our methods with other baselines. CW denotes the context window length, PE denotes the scaling method of RoPE, RC denotes reading comprehension, Common denotes commonsense question answering, Short Avg. denotes averaged scores on short benchmarks, and Avg. denotes averaged scores of all benchmarks. LongReD-C and LongReD-U denote our method with different skipped positional indices methods, \ie CREAM and Uniform Sampling.}
    \label{tab:results}
\end{table*}
\subsection{Experimental Settings}

\paratitle{Pre-training Setup.}
In the experiments, we choose Llama-3-8B~\cite{Dubey-arxiv-2023-llama3} and Mistral-7B-v0.3~\cite{jiang-arxiv-2023-mistral} to evaluate the effectiveness of our methods on context window extension. 
We sample the long text dataset $\mathcal{D}_1$ and short-to-long dataset $\mathcal{D}_3$ from SlimPajama following the setting of \citet{fu-icml-2024-data}. Inspired by~\citet{gao-arxiv-2024-how}, we employ higher-quality data as the short text dataset $\mathcal{D}_2$ with the length of 1K. Then, we trained our model on these three datasets with a token quantity ratio of 4:3:1 and a total of 1B tokens. The number of distillation layers $M$ is set to 3 for extending Llama-3-8B to 128K context window size and 6 for other settings for better balancing the short and long text modeling performances.
The details of training and parameter configurations and data mixture are displayed in Appendix~\ref{app:train_details}.


\paratitle{Evaluation Benchmarks.}
To thoroughly evaluate the short-text performance of long-context models, we choose 17 benchmarks covering 5 capacities, \ie general, coding, math, reading comprehension, and commonsense question answering. We also select RULER~\cite{Hsieh-arxiv-2024-RULER} to assess the long text processing abilities. The evaluation datasets and details are shown in Appendix~\ref{app:evaluation}.

\paratitle{Baselines.}
In our experiments, we choose two baseline methods for comparative analysis: (1) \emph{Long+CPT}: Only continual pre-training on long text datasets $\mathcal D_1$;
(2) \emph{Mix+CPT}: Continually pre-training on a mixed-length dataset ($\mathcal{D}_1\cup \mathcal{D}_2\cup\mathcal{D}_3$),  keeping the training data the same as our method. 


\subsection{Main Results}

Table~\ref{tab:results} presents the performance comparison between our proposed methods and various baselines on both short-text and long-text tasks. Detailed results for each dataset can be found in Appendix~\ref{app:all_results}.

First, LongReD performs better on short-text tasks and achieves competitive long-text modeling performances compared to baseline methods. Across various scaling methods and target context window sizes, LongReD consistently outperforms most baselines on short-text benchmarks. Specifically, when extending Llama-3-8B to a 32K-token context window using ABF and LongReD-C, the short-text capabilities are retained up to $99.4\%$ of the original model's performance, compared to $92.5\%$ achieved by naive long-text training.
Furthermore, our methods demonstrate competitive performance with Llama-3-8B and superior performance with Mistral-7B-v0.3 on the RULER benchmark for long-text processing, when compared to continual pre-training approaches. These compelling results robustly validate the efficacy of our proposed methodology.



Second, the skipped positional indices method is crucial for long-text performances. Comparing LongReD with CREAM and uniform sampling methods, we find that the short-text performances are similar while the long-text performances on RULER are largely different. When the extension ratio is relatively small (\eg, 4 times), CREAM demonstrates superior performance by effectively mitigating the lost-in-the-middle problem. However, on the length of 128K tokens, the model trained with CREAM performs largely worse than that with uniform sampling.  We hypothesize that the excessive focus on the middle positions introduced by CREAM results in some positions not being adequately trained.

Finally, our method benefits from the scaling techniques of positional encodings. When scaling LLaMA-3-8B to 32K with either ABF or PI, our LongReD consistently achieves superior short-text performances. However, a comparison between models with different extension methods reveals that PI exhibits inferior performance to ABF on both short-text and long-text tasks when combined with LongReD. 
We hypothesize that this performance gap arises due to the significant distribution discrepancy introduced by PI, which poses greater challenges for our method to mitigate effectively.

\subsection{Detailed Analysis}
\label{sec:ablation}
In this section, we conduct further detailed analysis to investigate our approach.


\paratitle{Ablation Study.}
In addition to long-text training, our method incorporates both short-text distillation and short-to-long distillation. To evaluate the individual contributions of these two components, we employ half of the training data for long-text training and the other half for short-text distillation or short-to-long distillation.
Moreover, we also evaluate the effect of the hyper-parameters $\alpha_1$ and 
$\alpha_2$ on both short and long text tasks. The results are compared against the LongReD-C method with $\alpha_1=5, \alpha_2 = 10$, as summarized in Table~\ref{tab:ablation}. The results indicate that excluding short-text distillation significantly degrades the model's performance on short-text tasks. Conversely, omitting the short-to-long distillation stage results in noticeable declines in long-text performance. In addition, the adaptation of the hyper-parameters is essential. Increasing or decreasing $\alpha_2$ will lead to performance degradation on long-text tasks, while slightly decreasing $\alpha_1$ will lead to better long-text performances. When either $\alpha_1$ or $\alpha_2$ is set to an extremely large value, the model's performance on long-context tasks drops precipitously. 



\begin{table}[htb]
    \centering
    \resizebox{\linewidth}{!}{
    \begin{tabular}{cc|cccccc}
         \toprule
        $\alpha_1$&$\alpha_2$ &General & Code&Math&Common&RC&RULER\\\midrule
         5&10&67.65 &39.90 &31.52& 72.35& 62.84 &84.98 \\\midrule
         -&15&66.54&39.93&31.68&70.64&62.47& 85.48\\
         5&-& 67.39 &41.25 &30.78 &71.50 &62.25&83.61\\\midrule
         5&100 & 66.96	&40.47	&33.12	&71.86	&62.47	&80.40\\
         5&30 & 67.00 & 40.14&31.59&72.26&62.29&83.82\\
         5&15&67.20&38.20&31.46&72.39&62.55&84.87 \\
         5&5&66.91& 41.84&32.67&72.36&62.42 & 83.75\\
         100&10& 65.33	&37.81&	29.24&	72.00&	61.85	&80.19\\
         2&10& 67.02&41.10 & 31.83& 72.25&63.04 & 86.03\\
         \bottomrule
    \end{tabular}}
    \caption{Ablation results on five tasks. ``-'' denotes the objective is not employed.}
    \label{tab:ablation}
\end{table}



\paratitle{Distillation Layers.} 
We explore the impact of distillation layers by varying the number of distilled layers and their selection strategies. In comparison to our method of selecting six layers via KL divergence (denoted as \emph{KL(6)}), we design three variants: (1) \emph{All}: distilling all layers; (2) \emph{Last}: distilling only the last layer; and (3) \emph{Uniform(6)}: uniformly selecting six layers. The results are summarized in Table~\ref{tab:distill_granularity}.
We can observe that distilling either all the layers or the last layer leads to worse long-text performance and general capacities than selecting six layers based on KL divergence. Furthermore, uniformly selected layers consistently exhibit inferior performance to those selected via attention KL divergence across the long-text tasks for the same number of distilled layers. These findings underscore the critical role of layer selection methods and distillation granularity in the performances of long-context LLMs.

       

\begin{table}[htb]
    \centering
    \resizebox{\linewidth}{!}{
    \begin{tabular}{c|cccccc}
         \toprule
         Layers &General & Code&Math&Common&RC&RULER\\\midrule
         KL(6)& 67.65 &39.90 &31.52& 72.35& 62.84 & 84.98 \\
         Uniform(6) &  66.86 & 40.24 & 32.19 & 72.33&60.27 & 82.53\\
         All&  66.89& 39.63 & 31.80& 72.33 & 62.61 & 81.47\\
         Last&   66.40 &  40.49 &30.69 &72.11 & 63.19& 82.24 \\\bottomrule
    \end{tabular}}
    \caption{Results with different distillation layers.}
    \label{tab:distill_granularity}
\end{table}

\paratitle{Distillation Length.} In our experiments, we set the length of short-text distillation $T_s$ as $1024$. To explore the effect of distillation length, we set $T_s$ as $2048$ and $8192$, and check the performance changes. As shown in Table~\ref{tab:distill_length}, increasing the distillation length will harm the model's ability to handle long texts while the short-text capacities are generally the same. 
Overfitting to the long distillation length will destroy the implicit positional information in hidden states, which may be the main reason for the performance drop (details are shown in Appendix~\ref{app:latent}).

\begin{table}[htb]
    \centering
    \resizebox{\linewidth}{!}{
    \begin{tabular}{c|cccccc}
         \toprule
         $T_s$&General & Code&Math&Common&RC&RULER\\\midrule
         1024& 67.65 &39.90 &31.52& 72.35& 62.84 & 84.98 \\
         2048& 66.80 &40.13 & 31.01 & 72.63& 62.32 & 83.20 \\
         8192& 67.43& 39.26 & 31.41 & 71.75 & 63.04 &  75.54 \\\bottomrule
    \end{tabular}}
    \caption{Results with varying distillation length $T_s$.}
    \label{tab:distill_length}
\end{table}





\paratitle{Comparison with Continual Learning Methods.} Beyond simple continual pre-training, we also compare LongReD with two continual learning methods that can effectively mitigate catastrophic forgetting issues: (1) \emph{Model Merging}: Following common model merging methods~\cite{hu-arxiv-2024-longrecipe,chen-arxiv-2024-extracting}, we average the checkpoints of the original model and the extended model; (2) \emph{Parameter-Efficient Tuning}: Since modifying RoPE directly affects attention computation without altering the knowledge stored in MLP parameters, we only tune the parameters in the attention blocks~\cite{chen-iclr-2024-longlora}. We train Llama-3-8B on a 32K context length with these methods and evaluate their performance. As shown in Table~\ref{tab:continual}, though these methods can achieve better performances than naive continual pre-training, our method consistently performs better on most short-context capacities evaluations. This may be because these methods only address catastrophic forgetting. However, our approach not only addresses catastrophic forgetting through continual learning but also leverages knowledge distillation to alleviate distribution drift.

\begin{table}[htb]
    \centering
    \resizebox{\linewidth}{!}{\begin{tabular}{c|cccccc}
        \toprule
         Method & General & Code & Math & Common & RC & RULER \\
        \midrule
          LongReD-C & 67.51 & 39.90 & 31.52 & 72.35 & 62.84 & 84.98 \\
          Merging & 65.92 & 35.04 & 32.38 & 71.61 & 62.65 & 83.38 \\
          PET & 66.82 & 39.20 & 30.16 & 70.78 & 62.26 & 85.86 \\
          CPT & 65.39 &29.32& 29.50 &69.83 & 60.98 & 82.80 \\        \bottomrule
    \end{tabular}}
    \caption{Comparison with continual learning methods, where Merging denotes model merging and PET denotes parameter-efficient tuning. }
    \label{tab:continual}
\end{table}


\section{Related Work}
\label{sec:related_work}
\paratitle{Context Window Extension.}
LLMs typically have constrained context windows owing to positional encodings and attention mechanisms~\cite{press-iclr-2022-alibi,han-arxiv-2023-lm,dong-arxiv-2023-survey}. In order to meet the growing demand for processing long texts~\cite{wang-emnlp-2024-rear, tang-arxiv-2025-unlocking,peng-arxiv-2025-cafe}, various methods have been proposed to extend the context windows of LLMs based on modification of RoPE~\cite{su-neurocomputing-2024-roformer}. Positional Interpolation down-scales positional indices to prevent out-of-distribution positions. In addition, NTK~\cite{bloc97-reddit-2023-ntk,emozilla-reddit-2023-dynamicntk}, ABF~\cite{xiong-naacl-2024-effective}, Yarn~\cite{peng-arxiv-2023-yarn}, and LongRoPE~\cite{ding-icml-2024-longrope} modify the base of RoPE to control the rotation angles of different dimensions. After modifying positional encodings, LLMs are lightweightly continued pre-training on long texts~\cite{fu-icml-2024-data}. Through this pipeline, the context windows of LLMs can be extended to 128K or even 1M tokens~\cite{Dubey-arxiv-2023-llama3,Zeng-arxiv-2024-glm}. Typically, they can achieve utilization of long texts within their context windows at the cost of degradation of general capacities on short texts~\cite{Dubey-arxiv-2023-llama3}. Our work mainly focuses on the reasons for the degradation of short-text capacities and approaches to mitigating the decline.

\paratitle{Knowledge Distillation.} Knowledge distillation serves as an approach to transferring knowledge and abilities of teacher models to student models~\cite{Hinton-arxiv-2015-Distilling,Xu-arxiv-2024-Survey}. During the era of LLMs, leveraging black-box and powerful LLMs, \eg GPT-4~\cite{openai-arxiv-2023-gpt4}, to generate instructions and even fine-grained reasoning path to training smaller LLMs is a common and effective method to enhance general and domain-specific capacities~\cite{vicuna2023,peng-arxiv-2023-instruction,mukherjee-arxiv-2023-orca,ho-acl-2023-large}. In addition, distilling the inner distribution of white-box LLMs can be another effective way. Divergence-based distillation methods minimize the divergence of the output probability distribution of teacher and student models~\cite{Gu-ICLR-2024-Minillm,agarwal-iclr-2024-on,jiang-arxiv-2024-MixCPT}. Similarity-based distillation methods align the intermediate hidden states, enabling the student models to process input like the teacher models~\cite{liang-icml-2023-less, Muralidharan-arxiv-2024-Compact}. Different from these methods of transferring knowledge of powerful models to small or compressed models, our method aims to restore and preserve the original capacities of LLMs during the context window extension stage.

\section{Conclusion}
\label{sec:conclusion}

In this paper, we analyzed two reasons for the performance degradation in short-text tasks after context window extension, \ie {distribution drift}  and {catastrophic forgetting}. Based on the observations, we proposed an effective training method to better preserve short-text capacities, named {Long} Context Pre-training with {Re}storation {D}istillation ({LongReD}). Besides continual pre-training on long texts, LongReD introduced two additional training objectives: short-text distillation and short-to-long distillation. Experiments demonstrated that our method could achieve better performances on common short-text benchmarks while achieving comparable long-text modeling abilities on the RULER benchmark compared with continual pre-training. In future work, we will explore how to directly integrate continual training with the distillation of the original model on long texts, rather than applying them separately to texts of different lengths.

\section*{Limitations}
\label{sec:limitation}

In this paper, we present a novel perspective on the extension of context windows and the associated decline in the general capabilities of LLMs. We also propose a method to preserve general capabilities while enhancing long-context abilities. However, a notable degradation in short-text performance after lightweight continual pretraining on only several billion tokens. \citet{xiong-naacl-2024-effective} and \citet{Dubey-arxiv-2023-llama3} have shown that, with training on more than 100 billion tokens, some capabilities on short texts of models may remain largely unaffected or even improve. Furthermore, our proposed method offers a general training strategy that is compatible with different positional encoding extension techniques. A more refined extension method that minimizes disruption to the model's distribution could further enhance the effectiveness of our approach, a topic we leave for future investigation.

\section*{Acknowledgement}

This work was partially supported by National Natural Science Foundation of China under Grant No.~92470205 and 62222215, Beijing Municipal Science and Technology Project under Grant No.~Z231100010323009, and Beijing Natural Science Foundation under Grant No. L233008. Xin Zhao is the corresponding author.

\bibliography{custom}

\begin{thebibliography}{79}
\providecommand{\natexlab}[1]{#1}

\bibitem[{Agarwal et~al.(2024)Agarwal, Vieillard, Zhou, Stanczyk, Garea, Geist, and Bachem}]{agarwal-iclr-2024-on}
Rishabh Agarwal, Nino Vieillard, Yongchao Zhou, Piotr Stanczyk, Sabela~Ramos Garea, Matthieu Geist, and Olivier Bachem. 2024.
\newblock On-policy distillation of language models: Learning from self-generated mistakes.
\newblock In \emph{The Twelfth International Conference on Learning Representations, {ICLR} 2024, Vienna, Austria, May 7-11, 2024}. OpenReview.net.

\bibitem[{Austin et~al.(2021)Austin, Odena, Nye, Bosma, Michalewski, Dohan, Jiang, Cai, Terry, Le, and Sutton}]{austin-arxiv-2021-mbpp}
Jacob Austin, Augustus Odena, Maxwell~I. Nye, Maarten Bosma, Henryk Michalewski, David Dohan, Ellen Jiang, Carrie~J. Cai, Michael Terry, Quoc~V. Le, and Charles Sutton. 2021.
\newblock Program synthesis with large language models.
\newblock \emph{CoRR}, abs/2108.07732.

\bibitem[{Azerbayev et~al.(2024)Azerbayev, Schoelkopf, Paster, Santos, McAleer, Jiang, Deng, Biderman, and Welleck}]{Azerbayev-ICLR-2024-proof}
Zhangir Azerbayev, Hailey Schoelkopf, Keiran Paster, Marco~Dos Santos, Stephen~Marcus McAleer, Albert~Q. Jiang, Jia Deng, Stella Biderman, and Sean Welleck. 2024.
\newblock Llemma: An open language model for mathematics.
\newblock In \emph{The Twelfth International Conference on Learning Representations, {ICLR} 2024, Vienna, Austria, May 7-11, 2024}. OpenReview.net.

\bibitem[{Bisk et~al.(2020)Bisk, Zellers, Bras, Gao, and Choi}]{bisk-aaai-2020-piqa}
Yonatan Bisk, Rowan Zellers, Ronan~Le Bras, Jianfeng Gao, and Yejin Choi. 2020.
\newblock {PIQA:} reasoning about physical commonsense in natural language.
\newblock In \emph{The Thirty-Fourth {AAAI} Conference on Artificial Intelligence, {AAAI} 2020, The Thirty-Second Innovative Applications of Artificial Intelligence Conference, {IAAI} 2020, The Tenth {AAAI} Symposium on Educational Advances in Artificial Intelligence, {EAAI} 2020, New York, NY, USA, February 7-12, 2020}, pages 7432--7439. {AAAI} Press.

\bibitem[{bloc97(2023)}]{bloc97-reddit-2023-ntk}
bloc97. 2023.
\newblock {NTK-Aware Scaled RoPE allows LLaMA models to have extended (8k+) context size without any fine-tuning and minimal perplexity degradation.}

\bibitem[{Brown et~al.(2020)Brown, Mann, Ryder, Subbiah, Kaplan, Dhariwal, Neelakantan, Shyam, Sastry, Askell, Agarwal, Herbert{-}Voss, Krueger, Henighan, Child, Ramesh, Ziegler, Wu, Winter, Hesse, Chen, Sigler, Litwin, Gray, Chess, Clark, Berner, McCandlish, Radford, Sutskever, and Amodei}]{brown-nips-2020-gpt3}
Tom~B. Brown, Benjamin Mann, Nick Ryder, Melanie Subbiah, Jared Kaplan, Prafulla Dhariwal, Arvind Neelakantan, Pranav Shyam, Girish Sastry, Amanda Askell, Sandhini Agarwal, Ariel Herbert{-}Voss, Gretchen Krueger, Tom Henighan, Rewon Child, Aditya Ramesh, Daniel~M. Ziegler, Jeffrey Wu, Clemens Winter, Christopher Hesse, Mark Chen, Eric Sigler, Mateusz Litwin, Scott Gray, Benjamin Chess, Jack Clark, Christopher Berner, Sam McCandlish, Alec Radford, Ilya Sutskever, and Dario Amodei. 2020.
\newblock Language models are few-shot learners.
\newblock In \emph{Advances in Neural Information Processing Systems 33: Annual Conference on Neural Information Processing Systems 2020, NeurIPS 2020, December 6-12, 2020, virtual}.

\bibitem[{Chen et~al.(2021)Chen, Tworek, Jun, Yuan, de~Oliveira~Pinto, Kaplan, Edwards, Burda, Joseph, Brockman, Ray, Puri, Krueger, Petrov, Khlaaf, Sastry, Mishkin, Chan, Gray, Ryder, Pavlov, Power, Kaiser, Bavarian, Winter, Tillet, Such, Cummings, Plappert, Chantzis, Barnes, Herbert{-}Voss, Guss, Nichol, Paino, Tezak, Tang, Babuschkin, Balaji, Jain, Saunders, Hesse, Carr, Leike, Achiam, Misra, Morikawa, Radford, Knight, Brundage, Murati, Mayer, Welinder, McGrew, Amodei, McCandlish, Sutskever, and Zaremba}]{chen-arxiv-2021-humaneval}
Mark Chen, Jerry Tworek, Heewoo Jun, Qiming Yuan, Henrique~Pond{\'{e}} de~Oliveira~Pinto, Jared Kaplan, Harri Edwards, Yuri Burda, Nicholas Joseph, Greg Brockman, Alex Ray, Raul Puri, Gretchen Krueger, Michael Petrov, Heidy Khlaaf, Girish Sastry, Pamela Mishkin, Brooke Chan, Scott Gray, Nick Ryder, Mikhail Pavlov, Alethea Power, Lukasz Kaiser, Mohammad Bavarian, Clemens Winter, Philippe Tillet, Felipe~Petroski Such, Dave Cummings, Matthias Plappert, Fotios Chantzis, Elizabeth Barnes, Ariel Herbert{-}Voss, William~Hebgen Guss, Alex Nichol, Alex Paino, Nikolas Tezak, Jie Tang, Igor Babuschkin, Suchir Balaji, Shantanu Jain, William Saunders, Christopher Hesse, Andrew~N. Carr, Jan Leike, Joshua Achiam, Vedant Misra, Evan Morikawa, Alec Radford, Matthew Knight, Miles Brundage, Mira Murati, Katie Mayer, Peter Welinder, Bob McGrew, Dario Amodei, Sam McCandlish, Ilya Sutskever, and Wojciech Zaremba. 2021.
\newblock Evaluating large language models trained on code.
\newblock \emph{CoRR}, abs/2107.03374.

\bibitem[{Chen et~al.(2023)Chen, Wong, Chen, and Tian}]{chen-arxiv-2023-extending}
Shouyuan Chen, Sherman Wong, Liangjian Chen, and Yuandong Tian. 2023.
\newblock Extending context window of large language models via positional interpolation.
\newblock \emph{CoRR}, abs/2306.15595.

\bibitem[{Chen et~al.(2024{\natexlab{a}})Chen, Qian, Tang, Lai, Liu, Han, and Jia}]{chen-iclr-2024-longlora}
Yukang Chen, Shengju Qian, Haotian Tang, Xin Lai, Zhijian Liu, Song Han, and Jiaya Jia. 2024{\natexlab{a}}.
\newblock Longlora: Efficient fine-tuning of long-context large language models.
\newblock In \emph{The Twelfth International Conference on Learning Representations, {ICLR} 2024, Vienna, Austria, May 7-11, 2024}. OpenReview.net.

\bibitem[{Chen et~al.(2025)Chen, Min, Zhang, Chen, Jiang, Cheng, Zhao, Liu, Miao, Lu, Fang, Wang, and Wen}]{chen-arxiv-2025-empirical}
Zhipeng Chen, Yingqian Min, Beichen Zhang, Jie Chen, Jinhao Jiang, Daixuan Cheng, Wayne~Xin Zhao, Zheng Liu, Xu~Miao, Yang Lu, Lei Fang, Zhongyuan Wang, and Ji{-}Rong Wen. 2025.
\newblock An empirical study on eliciting and improving r1-like reasoning models.
\newblock \emph{CoRR}, abs/2503.04548.

\bibitem[{Chen et~al.(2024{\natexlab{b}})Chen, Song, Zhou, Zhao, Wang, Chen, and Wen}]{chen-arxiv-2024-extracting}
Zhipeng Chen, Liang Song, Kun Zhou, Wayne~Xin Zhao, Bingning Wang, Weipeng Chen, and Ji{-}Rong Wen. 2024{\natexlab{b}}.
\newblock Extracting and transferring abilities for building multi-lingual ability-enhanced large language models.
\newblock \emph{CoRR}, abs/2410.07825.

\bibitem[{Chiang et~al.(2023)Chiang, Li, Lin, Sheng, Wu, Zhang, Zheng, Zhuang, Zhuang, Gonzalez, Stoica, and Xing}]{vicuna2023}
Wei-Lin Chiang, Zhuohan Li, Zi~Lin, Ying Sheng, Zhanghao Wu, Hao Zhang, Lianmin Zheng, Siyuan Zhuang, Yonghao Zhuang, Joseph~E. Gonzalez, Ion Stoica, and Eric~P. Xing. 2023.
\newblock \href {https://lmsys.org/blog/2023-03-30-vicuna/} {Vicuna: An open-source chatbot impressing gpt-4 with 90\%* chatgpt quality}.

\bibitem[{Choi et~al.(2018)Choi, He, Iyyer, Yatskar, Yih, Choi, Liang, and Zettlemoyer}]{choi-acl-2018-quac}
Eunsol Choi, He~He, Mohit Iyyer, Mark Yatskar, Wen{-}tau Yih, Yejin Choi, Percy Liang, and Luke Zettlemoyer. 2018.
\newblock Quac: Question answering in context.
\newblock In \emph{Proceedings of the 2018 Conference on Empirical Methods in Natural Language Processing, Brussels, Belgium, October 31 - November 4, 2018}, pages 2174--2184. Association for Computational Linguistics.

\bibitem[{Clark et~al.(2019)Clark, Lee, Chang, Kwiatkowski, Collins, and Toutanova}]{clark-naccl-2019-boolq}
Christopher Clark, Kenton Lee, Ming{-}Wei Chang, Tom Kwiatkowski, Michael Collins, and Kristina Toutanova. 2019.
\newblock Boolq: Exploring the surprising difficulty of natural yes/no questions.
\newblock In \emph{Proceedings of the 2019 Conference of the North American Chapter of the Association for Computational Linguistics: Human Language Technologies, {NAACL-HLT} 2019, Minneapolis, MN, USA, June 2-7, 2019, Volume 1 (Long and Short Papers)}, pages 2924--2936. Association for Computational Linguistics.

\bibitem[{Clark et~al.(2018)Clark, Cowhey, Etzioni, Khot, Sabharwal, Schoenick, and Tafjord}]{clark-arxiv-2018-arc}
Peter Clark, Isaac Cowhey, Oren Etzioni, Tushar Khot, Ashish Sabharwal, Carissa Schoenick, and Oyvind Tafjord. 2018.
\newblock Think you have solved question answering? try arc, the {AI2} reasoning challenge.
\newblock \emph{CoRR}, abs/1803.05457.

\bibitem[{Cobbe et~al.(2021)Cobbe, Kosaraju, Bavarian, Chen, Jun, Kaiser, Plappert, Tworek, Hilton, Nakano, Hesse, and Schulman}]{cobbe-arxiv-2021-gsm8k}
Karl Cobbe, Vineet Kosaraju, Mohammad Bavarian, Mark Chen, Heewoo Jun, Lukasz Kaiser, Matthias Plappert, Jerry Tworek, Jacob Hilton, Reiichiro Nakano, Christopher Hesse, and John Schulman. 2021.
\newblock Training verifiers to solve math word problems.
\newblock \emph{CoRR}, abs/2110.14168.

\bibitem[{Contributors(2023)}]{2023opencompass}
OpenCompass Contributors. 2023.
\newblock Opencompass: A universal evaluation platform for foundation models.
\newblock \url{https://github.com/open-compass/opencompass}.

\bibitem[{Ding et~al.(2024)Ding, Zhang, Zhang, Xu, Shang, Xu, Yang, and Yang}]{ding-icml-2024-longrope}
Yiran Ding, Li~Lyna Zhang, Chengruidong Zhang, Yuanyuan Xu, Ning Shang, Jiahang Xu, Fan Yang, and Mao Yang. 2024.
\newblock Longrope: Extending {LLM} context window beyond 2 million tokens.
\newblock In \emph{Forty-first International Conference on Machine Learning, {ICML} 2024, Vienna, Austria, July 21-27, 2024}. OpenReview.net.

\bibitem[{Dong et~al.(2024{\natexlab{a}})Dong, Li, Men, Zhao, Wang, Tian, Chen, and Wen}]{dong-arxiv-2024-exploring}
Zican Dong, Junyi Li, Xin Men, Wayne~Xin Zhao, Bingbing Wang, Zhen Tian, Weipeng Chen, and Ji{-}Rong Wen. 2024{\natexlab{a}}.
\newblock Exploring context window of large language models via decomposed positional vectors.
\newblock \emph{CoRR}, abs/2405.18009.

\bibitem[{Dong et~al.(2025)Dong, Peng, Liu, Zhao, Wu, Xiao, and Wang}]{dong-arxiv-2025-domain}
Zican Dong, Han Peng, Peiyu Liu, Wayne~Xin Zhao, Dong Wu, Feng Xiao, and Zhifeng Wang. 2025.
\newblock Domain-specific pruning of large mixture-of-experts models with few-shot demonstrations.
\newblock \emph{CoRR}, abs/2504.06792.

\bibitem[{Dong et~al.(2023)Dong, Tang, Li, and Zhao}]{dong-arxiv-2023-survey}
Zican Dong, Tianyi Tang, Junyi Li, and Wayne~Xin Zhao. 2023.
\newblock A survey on long text modeling with transformers.
\newblock \emph{CoRR}, abs/2302.14502.

\bibitem[{Dong et~al.(2024{\natexlab{b}})Dong, Tang, Li, Zhao, and Wen}]{Dong-COLING-2024-BAMBOO}
Zican Dong, Tianyi Tang, Junyi Li, Wayne~Xin Zhao, and Ji{-}Rong Wen. 2024{\natexlab{b}}.
\newblock {BAMBOO:} {A} comprehensive benchmark for evaluating long text modeling capacities of large language models.
\newblock In \emph{Proceedings of the 2024 Joint International Conference on Computational Linguistics, Language Resources and Evaluation, {LREC/COLING} 2024, 20-25 May, 2024, Torino, Italy}, pages 2086--2099. {ELRA} and {ICCL}.

\bibitem[{Dua et~al.(2019)Dua, Wang, Dasigi, Stanovsky, Singh, and Gardner}]{dua-naacl-2019-drop}
Dheeru Dua, Yizhong Wang, Pradeep Dasigi, Gabriel Stanovsky, Sameer Singh, and Matt Gardner. 2019.
\newblock {DROP:} {A} reading comprehension benchmark requiring discrete reasoning over paragraphs.
\newblock In \emph{Proceedings of the 2019 Conference of the North American Chapter of the Association for Computational Linguistics: Human Language Technologies, {NAACL-HLT} 2019, Minneapolis, MN, USA, June 2-7, 2019, Volume 1 (Long and Short Papers)}, pages 2368--2378. Association for Computational Linguistics.

\bibitem[{Dubey et~al.(2024)Dubey, Jauhri, Pandey, Kadian, Al{-}Dahle, Letman, Mathur, Schelten, Yang, Fan, Goyal, Hartshorn, Yang, Mitra, Sravankumar, Korenev, Hinsvark, Rao, Zhang, Rodriguez, Gregerson, Spataru, Rozi{\`{e}}re, Biron, Tang, Chern, Caucheteux, Nayak, Bi, Marra, McConnell, Keller, Touret, Wu, Wong, Ferrer, Nikolaidis, Allonsius, Song, Pintz, Livshits, Esiobu, Choudhary, Mahajan, Garcia{-}Olano, Perino, Hupkes, Lakomkin, AlBadawy, Lobanova, Dinan, Smith, Radenovic, Zhang, Synnaeve, Lee, Anderson, Nail, Mialon, Pang, Cucurell, Nguyen, Korevaar, Xu, Touvron, Zarov, Ibarra, Kloumann, Misra, Evtimov, Copet, Lee, Geffert, Vranes, Park, Mahadeokar, Shah, van~der Linde, Billock, Hong, Lee, Fu, Chi, Huang, Liu, Wang, Yu, Bitton, Spisak, Park, Rocca, Johnstun, Saxe, Jia, Alwala, Upasani, Plawiak, Li, Heafield, Stone, and et~al.}]{Dubey-arxiv-2023-llama3}
Abhimanyu Dubey, Abhinav Jauhri, Abhinav Pandey, Abhishek Kadian, Ahmad Al{-}Dahle, Aiesha Letman, Akhil Mathur, Alan Schelten, Amy Yang, Angela Fan, Anirudh Goyal, Anthony Hartshorn, Aobo Yang, Archi Mitra, Archie Sravankumar, Artem Korenev, Arthur Hinsvark, Arun Rao, Aston Zhang, Aur{\'{e}}lien Rodriguez, Austen Gregerson, Ava Spataru, Baptiste Rozi{\`{e}}re, Bethany Biron, Binh Tang, Bobbie Chern, Charlotte Caucheteux, Chaya Nayak, Chloe Bi, Chris Marra, Chris McConnell, Christian Keller, Christophe Touret, Chunyang Wu, Corinne Wong, Cristian~Canton Ferrer, Cyrus Nikolaidis, Damien Allonsius, Daniel Song, Danielle Pintz, Danny Livshits, David Esiobu, Dhruv Choudhary, Dhruv Mahajan, Diego Garcia{-}Olano, Diego Perino, Dieuwke Hupkes, Egor Lakomkin, Ehab AlBadawy, Elina Lobanova, Emily Dinan, Eric~Michael Smith, Filip Radenovic, Frank Zhang, Gabriel Synnaeve, Gabrielle Lee, Georgia~Lewis Anderson, Graeme Nail, Gr{\'{e}}goire Mialon, Guan Pang, Guillem Cucurell, Hailey Nguyen, Hannah Korevaar, Hu~Xu, Hugo
  Touvron, Iliyan Zarov, Imanol~Arrieta Ibarra, Isabel~M. Kloumann, Ishan Misra, Ivan Evtimov, Jade Copet, Jaewon Lee, Jan Geffert, Jana Vranes, Jason Park, Jay Mahadeokar, Jeet Shah, Jelmer van~der Linde, Jennifer Billock, Jenny Hong, Jenya Lee, Jeremy Fu, Jianfeng Chi, Jianyu Huang, Jiawen Liu, Jie Wang, Jiecao Yu, Joanna Bitton, Joe Spisak, Jongsoo Park, Joseph Rocca, Joshua Johnstun, Joshua Saxe, Junteng Jia, Kalyan~Vasuden Alwala, Kartikeya Upasani, Kate Plawiak, Ke~Li, Kenneth Heafield, Kevin Stone, and et~al. 2024.
\newblock The llama 3 herd of models.
\newblock \emph{CoRR}, abs/2407.21783.

\bibitem[{emozilla(2023)}]{emozilla-reddit-2023-dynamicntk}
emozilla. 2023.
\newblock {Dynamically Scaled RoPE further increases performance of long context LLaMA with zero fine-tuning}.

\bibitem[{Fu et~al.(2024)Fu, Panda, Niu, Yue, Hajishirzi, Kim, and Peng}]{fu-icml-2024-data}
Yao Fu, Rameswar Panda, Xinyao Niu, Xiang Yue, Hannaneh Hajishirzi, Yoon Kim, and Hao Peng. 2024.
\newblock Data engineering for scaling language models to 128k context.
\newblock In \emph{Forty-first International Conference on Machine Learning, {ICML} 2024, Vienna, Austria, July 21-27, 2024}. OpenReview.net.

\bibitem[{Gao et~al.(2024)Gao, Wettig, Yen, and Chen}]{gao-arxiv-2024-how}
Tianyu Gao, Alexander Wettig, Howard Yen, and Danqi Chen. 2024.
\newblock How to train long-context language models (effectively).
\newblock \emph{CoRR}, abs/2410.02660.

\bibitem[{Gu et~al.(2024)Gu, Dong, Wei, and Huang}]{Gu-ICLR-2024-Minillm}
Yuxian Gu, Li~Dong, Furu Wei, and Minlie Huang. 2024.
\newblock Minillm: Knowledge distillation of large language models.
\newblock In \emph{The Twelfth International Conference on Learning Representations, {ICLR} 2024, Vienna, Austria, May 7-11, 2024}. OpenReview.net.

\bibitem[{Han et~al.(2023)Han, Wang, Xiong, Chen, Ji, and Wang}]{han-arxiv-2023-lm}
Chi Han, Qifan Wang, Wenhan Xiong, Yu~Chen, Heng Ji, and Sinong Wang. 2023.
\newblock Lm-infinite: Simple on-the-fly length generalization for large language models.
\newblock \emph{CoRR}, abs/2308.16137.

\bibitem[{Hendrycks et~al.(2021{\natexlab{a}})Hendrycks, Burns, Basart, Zou, Mazeika, Song, and Steinhardt}]{Hendrycks-iclr-2021-mmlu}
Dan Hendrycks, Collin Burns, Steven Basart, Andy Zou, Mantas Mazeika, Dawn Song, and Jacob Steinhardt. 2021{\natexlab{a}}.
\newblock Measuring massive multitask language understanding.
\newblock In \emph{9th International Conference on Learning Representations, {ICLR} 2021, Virtual Event, Austria, May 3-7, 2021}. OpenReview.net.

\bibitem[{Hendrycks et~al.(2021{\natexlab{b}})Hendrycks, Burns, Kadavath, Arora, Basart, Tang, Song, and Steinhardt}]{henderycksnips-2021-math}
Dan Hendrycks, Collin Burns, Saurav Kadavath, Akul Arora, Steven Basart, Eric Tang, Dawn Song, and Jacob Steinhardt. 2021{\natexlab{b}}.
\newblock Measuring mathematical problem solving with the {MATH} dataset.
\newblock In \emph{Proceedings of the Neural Information Processing Systems Track on Datasets and Benchmarks 1, NeurIPS Datasets and Benchmarks 2021, December 2021, virtual}.

\bibitem[{Hinton et~al.(2015)Hinton, Vinyals, and Dean}]{Hinton-arxiv-2015-Distilling}
Geoffrey~E. Hinton, Oriol Vinyals, and Jeffrey Dean. 2015.
\newblock Distilling the knowledge in a neural network.
\newblock \emph{CoRR}, abs/1503.02531.

\bibitem[{Ho et~al.(2023)Ho, Schmid, and Yun}]{ho-acl-2023-large}
Namgyu Ho, Laura Schmid, and Se{-}Young Yun. 2023.
\newblock Large language models are reasoning teachers.
\newblock In \emph{Proceedings of the 61st Annual Meeting of the Association for Computational Linguistics (Volume 1: Long Papers), {ACL} 2023, Toronto, Canada, July 9-14, 2023}, pages 14852--14882. Association for Computational Linguistics.

\bibitem[{Hsieh et~al.(2024)Hsieh, Sun, Kriman, Acharya, Rekesh, Jia, Zhang, and Ginsburg}]{Hsieh-arxiv-2024-RULER}
Cheng{-}Ping Hsieh, Simeng Sun, Samuel Kriman, Shantanu Acharya, Dima Rekesh, Fei Jia, Yang Zhang, and Boris Ginsburg. 2024.
\newblock {RULER:} what's the real context size of your long-context language models?
\newblock \emph{CoRR}, abs/2404.06654.

\bibitem[{Hu et~al.(2024{\natexlab{a}})Hu, Tu, Han, He, Cui, Long, Zheng, Fang, Huang, Zhao, Zhang, Thai, Zhang, Wang, Yao, Zhao, Zhou, Cai, Zhai, Ding, Jia, Zeng, Li, Liu, and Sun}]{hu-arxiv-2024-minicpm}
Shengding Hu, Yuge Tu, Xu~Han, Chaoqun He, Ganqu Cui, Xiang Long, Zhi Zheng, Yewei Fang, Yuxiang Huang, Weilin Zhao, Xinrong Zhang, Zhen~Leng Thai, Kai Zhang, Chongyi Wang, Yuan Yao, Chenyang Zhao, Jie Zhou, Jie Cai, Zhongwu Zhai, Ning Ding, Chao Jia, Guoyang Zeng, Dahai Li, Zhiyuan Liu, and Maosong Sun. 2024{\natexlab{a}}.
\newblock Minicpm: Unveiling the potential of small language models with scalable training strategies.
\newblock \emph{CoRR}, abs/2404.06395.

\bibitem[{Hu et~al.(2024{\natexlab{b}})Hu, Liu, Zhao, Wang, Wang, Shen, Gu, Luu, Ng, Jiang et~al.}]{hu-arxiv-2024-longrecipe}
Zhiyuan Hu, Yuliang Liu, Jinman Zhao, Suyuchen Wang, Yan Wang, Wei Shen, Qing Gu, Anh~Tuan Luu, See-Kiong Ng, Zhiwei Jiang, et~al. 2024{\natexlab{b}}.
\newblock Longrecipe: Recipe for efficient long context generalization in large language models.
\newblock \emph{CoRR}, abs/2409.00509.

\bibitem[{Jiang et~al.(2023)Jiang, Sablayrolles, Mensch, Bamford, Chaplot, de~Las~Casas, Bressand, Lengyel, Lample, Saulnier, Lavaud, Lachaux, Stock, Scao, Lavril, Wang, Lacroix, and Sayed}]{jiang-arxiv-2023-mistral}
Albert~Q. Jiang, Alexandre Sablayrolles, Arthur Mensch, Chris Bamford, Devendra~Singh Chaplot, Diego de~Las~Casas, Florian Bressand, Gianna Lengyel, Guillaume Lample, Lucile Saulnier, L{\'{e}}lio~Renard Lavaud, Marie{-}Anne Lachaux, Pierre Stock, Teven~Le Scao, Thibaut Lavril, Thomas Wang, Timoth{\'{e}}e Lacroix, and William~El Sayed. 2023.
\newblock Mistral 7b.
\newblock \emph{CoRR}, abs/2310.06825.

\bibitem[{Jiang et~al.(2024)Jiang, Li, Zhao, Song, Zhang, and Wen}]{jiang-arxiv-2024-MixCPT}
Jinhao Jiang, Junyi Li, Wayne~Xin Zhao, Yang Song, Tao Zhang, and Ji{-}Rong Wen. 2024.
\newblock Mix-cpt: {A} domain adaptation framework via decoupling knowledge learning and format alignment.
\newblock \emph{CoRR}, abs/2407.10804.

\bibitem[{Joshi et~al.(2017)Joshi, Choi, Weld, and Zettlemoyer}]{Joshi-acl-2017-triviaqa}
Mandar Joshi, Eunsol Choi, Daniel~S. Weld, and Luke Zettlemoyer. 2017.
\newblock Triviaqa: {A} large scale distantly supervised challenge dataset for reading comprehension.
\newblock In \emph{Proceedings of the 55th Annual Meeting of the Association for Computational Linguistics, {ACL} 2017, Vancouver, Canada, July 30 - August 4, Volume 1: Long Papers}, pages 1601--1611. Association for Computational Linguistics.

\bibitem[{Liang et~al.(2023)Liang, Zuo, Zhang, He, Chen, and Zhao}]{liang-icml-2023-less}
Chen Liang, Simiao Zuo, Qingru Zhang, Pengcheng He, Weizhu Chen, and Tuo Zhao. 2023.
\newblock Less is more: Task-aware layer-wise distillation for language model compression.
\newblock In \emph{International Conference on Machine Learning, {ICML} 2023, 23-29 July 2023, Honolulu, Hawaii, {USA}}, volume 202 of \emph{Proceedings of Machine Learning Research}, pages 20852--20867. {PMLR}.

\bibitem[{Liu et~al.(2023)Liu, Zaharia, and Abbeel}]{liu-arxiv-2023-ringattn}
Hao Liu, Matei Zaharia, and Pieter Abbeel. 2023.
\newblock Ring attention with blockwise transformers for near-infinite context.
\newblock \emph{CoRR}, abs/2310.01889.

\bibitem[{Loshchilov and Hutter(2019)}]{Loshchilov-iclr-2019-adamw}
Ilya Loshchilov and Frank Hutter. 2019.
\newblock Decoupled weight decay regularization.
\newblock In \emph{7th International Conference on Learning Representations, {ICLR} 2019, New Orleans, LA, USA, May 6-9, 2019}. OpenReview.net.

\bibitem[{Lozhkov et~al.(2024)Lozhkov, Li, Allal, Cassano, Lamy{-}Poirier, Tazi, Tang, Pykhtar, Liu, Wei, Liu, Tian, Kocetkov, Zucker, Belkada, Wang, Liu, Abulkhanov, Paul, Li, Li, Risdal, Li, Zhu, Zhuo, Zheltonozhskii, Dade, Yu, Krau{\ss}, Jain, Su, He, Dey, Abati, Chai, Muennighoff, Tang, Oblokulov, Akiki, Marone, Mou, Mishra, Gu, Hui, Dao, Zebaze, Dehaene, Patry, Xu, McAuley, Hu, Scholak, Paquet, Robinson, Anderson, Chapados, and et~al.}]{Lozhkov-arxiv-2024-stack}
Anton Lozhkov, Raymond Li, Loubna~Ben Allal, Federico Cassano, Joel Lamy{-}Poirier, Nouamane Tazi, Ao~Tang, Dmytro Pykhtar, Jiawei Liu, Yuxiang Wei, Tianyang Liu, Max Tian, Denis Kocetkov, Arthur Zucker, Younes Belkada, Zijian Wang, Qian Liu, Dmitry Abulkhanov, Indraneil Paul, Zhuang Li, Wen{-}Ding Li, Megan Risdal, Jia Li, Jian Zhu, Terry~Yue Zhuo, Evgenii Zheltonozhskii, Nii Osae~Osae Dade, Wenhao Yu, Lucas Krau{\ss}, Naman Jain, Yixuan Su, Xuanli He, Manan Dey, Edoardo Abati, Yekun Chai, Niklas Muennighoff, Xiangru Tang, Muhtasham Oblokulov, Christopher Akiki, Marc Marone, Chenghao Mou, Mayank Mishra, Alex Gu, Binyuan Hui, Tri Dao, Armel Zebaze, Olivier Dehaene, Nicolas Patry, Canwen Xu, Julian~J. McAuley, Han Hu, Torsten Scholak, S{\'{e}}bastien Paquet, Jennifer Robinson, Carolyn~Jane Anderson, Nicolas Chapados, and et~al. 2024.
\newblock Starcoder 2 and the stack v2: The next generation.
\newblock \emph{CoRR}, abs/2402.19173.

\bibitem[{Men et~al.(2024)Men, Xu, Zhang, Wang, Lin, Lu, Han, and Chen}]{men-arxiv-2024-shortgpt}
Xin Men, Mingyu Xu, Qingyu Zhang, Bingning Wang, Hongyu Lin, Yaojie Lu, Xianpei Han, and Weipeng Chen. 2024.
\newblock Shortgpt: Layers in large language models are more redundant than you expect.
\newblock \emph{CoRR}, abs/2403.03853.

\bibitem[{Mihaylov et~al.(2018)Mihaylov, Clark, Khot, and Sabharwal}]{todor-emnlp-2018-openbookqa}
Todor Mihaylov, Peter Clark, Tushar Khot, and Ashish Sabharwal. 2018.
\newblock Can a suit of armor conduct electricity? {A} new dataset for open book question answering.
\newblock In \emph{Proceedings of the 2018 Conference on Empirical Methods in Natural Language Processing, Brussels, Belgium, October 31 - November 4, 2018}, pages 2381--2391. Association for Computational Linguistics.

\bibitem[{Mukherjee et~al.(2023)Mukherjee, Mitra, Jawahar, Agarwal, Palangi, and Awadallah}]{mukherjee-arxiv-2023-orca}
Subhabrata Mukherjee, Arindam Mitra, Ganesh Jawahar, Sahaj Agarwal, Hamid Palangi, and Ahmed Awadallah. 2023.
\newblock Orca: Progressive learning from complex explanation traces of {GPT-4}.
\newblock \emph{CoRR}, abs/2306.02707.

\bibitem[{Muralidharan et~al.(2024)Muralidharan, Sreenivas, Joshi, Chochowski, Patwary, Shoeybi, Catanzaro, Kautz, and Molchanov}]{Muralidharan-arxiv-2024-Compact}
Saurav Muralidharan, Sharath~Turuvekere Sreenivas, Raviraj Joshi, Marcin Chochowski, Mostofa Patwary, Mohammad Shoeybi, Bryan Catanzaro, Jan Kautz, and Pavlo Molchanov. 2024.
\newblock Compact language models via pruning and knowledge distillation.
\newblock \emph{CoRR}, abs/2407.14679.

\bibitem[{OpenAI(2023)}]{openai-arxiv-2023-gpt4}
OpenAI. 2023.
\newblock {GPT-4} technical report.
\newblock \emph{CoRR}, abs/2303.08774.

\bibitem[{Paperno et~al.(2016)Paperno, Kruszewski, Lazaridou, Pham, Bernardi, Pezzelle, Baroni, Boleda, and Fern{\'{a}}ndez}]{paperno-acl-2016-lambada}
Denis Paperno, Germ{\'{a}}n Kruszewski, Angeliki Lazaridou, Quan~Ngoc Pham, Raffaella Bernardi, Sandro Pezzelle, Marco Baroni, Gemma Boleda, and Raquel Fern{\'{a}}ndez. 2016.
\newblock The {LAMBADA} dataset: Word prediction requiring a broad discourse context.
\newblock In \emph{Proceedings of the 54th Annual Meeting of the Association for Computational Linguistics, {ACL} 2016, August 7-12, 2016, Berlin, Germany, Volume 1: Long Papers}. The Association for Computer Linguistics.

\bibitem[{Pekelis et~al.(2024)Pekelis, Feil, Moret, Huang, and Peng}]{gradient-hf-2024-longcontextllama3}
Leonid Pekelis, Michael Feil, Forrest Moret, Mark Huang, and Tiffany Peng. 2024.
\newblock \href {https://doi.org/10.57967/hf/3372} {Llama 3 gradient: A series of long context models}.

\bibitem[{Penedo et~al.(2024)Penedo, Kydl{\'{\i}}cek, Allal, Lozhkov, Mitchell, Raffel, von Werra, and Wolf}]{penedo-arxiv-2024-fineweb}
Guilherme Penedo, Hynek Kydl{\'{\i}}cek, Loubna~Ben Allal, Anton Lozhkov, Margaret Mitchell, Colin Raffel, Leandro von Werra, and Thomas Wolf. 2024.
\newblock The fineweb datasets: Decanting the web for the finest text data at scale.
\newblock \emph{CoRR}, abs/2406.17557.

\bibitem[{Peng et~al.(2023{\natexlab{a}})Peng, Li, He, Galley, and Gao}]{peng-arxiv-2023-instruction}
Baolin Peng, Chunyuan Li, Pengcheng He, Michel Galley, and Jianfeng Gao. 2023{\natexlab{a}}.
\newblock Instruction tuning with {GPT-4}.
\newblock \emph{CoRR}, abs/2304.03277.

\bibitem[{Peng et~al.(2023{\natexlab{b}})Peng, Quesnelle, Fan, and Shippole}]{peng-arxiv-2023-yarn}
Bowen Peng, Jeffrey Quesnelle, Honglu Fan, and Enrico Shippole. 2023{\natexlab{b}}.
\newblock Yarn: Efficient context window extension of large language models.
\newblock \emph{CoRR}, abs/2309.00071.

\bibitem[{Peng et~al.(2025)Peng, Jiang, Dong, Zhao, and Fang}]{peng-arxiv-2025-cafe}
Han Peng, Jinhao Jiang, Zican Dong, Wayne~Xin Zhao, and Lei Fang. 2025.
\newblock Cafe: Retrieval head-based coarse-to-fine information seeking to enhance multi-document qa capability.
\newblock \emph{arXiv preprint arXiv:2505.10063}.

\bibitem[{Press et~al.(2022)Press, Smith, and Lewis}]{press-iclr-2022-alibi}
Ofir Press, Noah~A. Smith, and Mike Lewis. 2022.
\newblock Train short, test long: Attention with linear biases enables input length extrapolation.
\newblock In \emph{The Tenth International Conference on Learning Representations, {ICLR} 2022, Virtual Event, April 25-29, 2022}. OpenReview.net.

\bibitem[{Rae et~al.(2020)Rae, Potapenko, Jayakumar, Hillier, and Lillicrap}]{rae-iclr-2020-pg19}
Jack~W. Rae, Anna Potapenko, Siddhant~M. Jayakumar, Chloe Hillier, and Timothy~P. Lillicrap. 2020.
\newblock Compressive transformers for long-range sequence modelling.
\newblock In \emph{8th International Conference on Learning Representations, {ICLR} 2020, Addis Ababa, Ethiopia, April 26-30, 2020}. OpenReview.net.

\bibitem[{Rajpurkar et~al.(2016)Rajpurkar, Zhang, Lopyrev, and Liang}]{rajpurkar-EMNLP-2016-squad}
Pranav Rajpurkar, Jian Zhang, Konstantin Lopyrev, and Percy Liang. 2016.
\newblock Squad: 100, 000+ questions for machine comprehension of text.
\newblock In \emph{Proceedings of the 2016 Conference on Empirical Methods in Natural Language Processing, {EMNLP} 2016, Austin, Texas, USA, November 1-4, 2016}, pages 2383--2392. The Association for Computational Linguistics.

\bibitem[{Sap et~al.(2019)Sap, Rashkin, Chen, Bras, and Choi}]{Sap-emnlp-2019-siqa}
Maarten Sap, Hannah Rashkin, Derek Chen, Ronan~Le Bras, and Yejin Choi. 2019.
\newblock Social iqa: Commonsense reasoning about social interactions.
\newblock In \emph{Proceedings of the 2019 Conference on Empirical Methods in Natural Language Processing and the 9th International Joint Conference on Natural Language Processing, {EMNLP-IJCNLP} 2019, Hong Kong, China, November 3-7, 2019}, pages 4462--4472. Association for Computational Linguistics.

\bibitem[{Soboleva et~al.(2023)Soboleva, Al-Khateeb, Myers, Steeves, Hestness, and Dey}]{soboleva-huggingface-2023-slimpajama}
Daria Soboleva, Faisal Al-Khateeb, Robert Myers, Jacob~R Steeves, Joel Hestness, and Nolan Dey. 2023.
\newblock {SlimPajama: A 627B token cleaned and deduplicated version of RedPajama}.

\bibitem[{Su et~al.(2024)Su, Ahmed, Lu, Pan, Bo, and Liu}]{su-neurocomputing-2024-roformer}
Jianlin Su, Murtadha H.~M. Ahmed, Yu~Lu, Shengfeng Pan, Wen Bo, and Yunfeng Liu. 2024.
\newblock Roformer: Enhanced transformer with rotary position embedding.
\newblock \emph{Neurocomputing}, 568:127063.

\bibitem[{Suzgun et~al.(2023)Suzgun, Scales, Sch{\"{a}}rli, Gehrmann, Tay, Chung, Chowdhery, Le, Chi, Zhou, and Wei}]{suzgun-acl-2023-bbh}
Mirac Suzgun, Nathan Scales, Nathanael Sch{\"{a}}rli, Sebastian Gehrmann, Yi~Tay, Hyung~Won Chung, Aakanksha Chowdhery, Quoc~V. Le, Ed~H. Chi, Denny Zhou, and Jason Wei. 2023.
\newblock Challenging big-bench tasks and whether chain-of-thought can solve them.
\newblock In \emph{Findings of the Association for Computational Linguistics: {ACL} 2023, Toronto, Canada, July 9-14, 2023}, pages 13003--13051. Association for Computational Linguistics.

\bibitem[{Talmor et~al.(2019)Talmor, Herzig, Lourie, and Berant}]{talmor-naacl-2019-commonsenseqa}
Alon Talmor, Jonathan Herzig, Nicholas Lourie, and Jonathan Berant. 2019.
\newblock Commonsenseqa: {A} question answering challenge targeting commonsense knowledge.
\newblock In \emph{Proceedings of the 2019 Conference of the North American Chapter of the Association for Computational Linguistics: Human Language Technologies, {NAACL-HLT} 2019, Minneapolis, MN, USA, June 2-7, 2019, Volume 1 (Long and Short Papers)}, pages 4149--4158. Association for Computational Linguistics.

\bibitem[{Tang et~al.(2024)Tang, Hu, Li, Luo, Qin, Sun, Wang, Xu, Cheng, Guo, Peng, Zheng, Tang, Min, Chen, Chen, Zhao, Ding, Wang, Dong, Xia, Li, Zhou, Zhao, and Wen}]{tang-arxiv-2024-llmbox}
Tianyi Tang, Yiwen Hu, Bingqian Li, Wenyang Luo, Zijing Qin, Haoxiang Sun, Jiapeng Wang, Shiyi Xu, Xiaoxue Cheng, Geyang Guo, Han Peng, Bowen Zheng, Yiru Tang, Yingqian Min, Yushuo Chen, Jie Chen, Yuanqian Zhao, Luran Ding, Yuhao Wang, Zican Dong, Chunxuan Xia, Junyi Li, Kun Zhou, Wayne~Xin Zhao, and Ji{-}Rong Wen. 2024.
\newblock Llmbox: {A} comprehensive library for large language models.
\newblock \emph{CoRR}, abs/2407.05563.

\bibitem[{Tang et~al.(2025{\natexlab{a}})Tang, Wang, Lv, Min, Zhao, Hu, Liu, and Zhang}]{tang-arxiv-2025-unlocking}
Xinyu Tang, Xiaolei Wang, Zhihao Lv, Yingqian Min, Wayne~Xin Zhao, Binbin Hu, Ziqi Liu, and Zhiqiang Zhang. 2025{\natexlab{a}}.
\newblock Unlocking general long chain-of-thought reasoning capabilities of large language models via representation engineering.
\newblock \emph{CoRR}, abs/2503.11314.

\bibitem[{Tang et~al.(2025{\natexlab{b}})Tang, Wang, Zhao, Lu, Li, and Wen}]{tang-aaai-2025-unleashing}
Xinyu Tang, Xiaolei Wang, Wayne~Xin Zhao, Siyuan Lu, Yaliang Li, and Ji{-}Rong Wen. 2025{\natexlab{b}}.
\newblock Unleashing the potential of large language models as prompt optimizers: Analogical analysis with gradient-based model optimizers.
\newblock In \emph{AAAI-25, Sponsored by the Association for the Advancement of Artificial Intelligence, February 25 - March 4, 2025, Philadelphia, PA, {USA}}, pages 25264--25272. {AAAI} Press.

\bibitem[{Tang et~al.(2025{\natexlab{c}})Tang, Wang, Zhao, and Wen}]{tang-naacl-2025-dawnicl}
Xinyu Tang, Xiaolei Wang, Xin Zhao, and Ji{-}Rong Wen. 2025{\natexlab{c}}.
\newblock {DAWN-ICL:} strategic planning of problem-solving trajectories for zero-shot in-context learning.
\newblock In \emph{Proceedings of the 2025 Conference of the Nations of the Americas Chapter of the Association for Computational Linguistics: Human Language Technologies, {NAACL} 2025 - Volume 1: Long Papers, Albuquerque, New Mexico, USA, April 29 - May 4, 2025}, pages 1918--1934. Association for Computational Linguistics.

\bibitem[{Touvron et~al.(2023)Touvron, Martin, Stone, Albert, Almahairi, Babaei, Bashlykov, Batra, Bhargava, Bhosale, Bikel, Blecher, Canton{-}Ferrer, Chen, Cucurull, Esiobu, Fernandes, Fu, Fu, Fuller, Gao, Goswami, Goyal, Hartshorn, Hosseini, Hou, Inan, Kardas, Kerkez, Khabsa, Kloumann, Korenev, Koura, Lachaux, Lavril, Lee, Liskovich, Lu, Mao, Martinet, Mihaylov, Mishra, Molybog, Nie, Poulton, Reizenstein, Rungta, Saladi, Schelten, Silva, Smith, Subramanian, Tan, Tang, Taylor, Williams, Kuan, Xu, Yan, Zarov, Zhang, Fan, Kambadur, Narang, Rodriguez, Stojnic, Edunov, and Scialom}]{Touvron-arxiv-2023-llama2}
Hugo Touvron, Louis Martin, Kevin Stone, Peter Albert, Amjad Almahairi, Yasmine Babaei, Nikolay Bashlykov, Soumya Batra, Prajjwal Bhargava, Shruti Bhosale, Dan Bikel, Lukas Blecher, Cristian Canton{-}Ferrer, Moya Chen, Guillem Cucurull, David Esiobu, Jude Fernandes, Jeremy Fu, Wenyin Fu, Brian Fuller, Cynthia Gao, Vedanuj Goswami, Naman Goyal, Anthony Hartshorn, Saghar Hosseini, Rui Hou, Hakan Inan, Marcin Kardas, Viktor Kerkez, Madian Khabsa, Isabel Kloumann, Artem Korenev, Punit~Singh Koura, Marie{-}Anne Lachaux, Thibaut Lavril, Jenya Lee, Diana Liskovich, Yinghai Lu, Yuning Mao, Xavier Martinet, Todor Mihaylov, Pushkar Mishra, Igor Molybog, Yixin Nie, Andrew Poulton, Jeremy Reizenstein, Rashi Rungta, Kalyan Saladi, Alan Schelten, Ruan Silva, Eric~Michael Smith, Ranjan Subramanian, Xiaoqing~Ellen Tan, Binh Tang, Ross Taylor, Adina Williams, Jian~Xiang Kuan, Puxin Xu, Zheng Yan, Iliyan Zarov, Yuchen Zhang, Angela Fan, Melanie Kambadur, Sharan Narang, Aur{\'{e}}lien Rodriguez, Robert Stojnic, Sergey Edunov,
  and Thomas Scialom. 2023.
\newblock Llama 2: Open foundation and fine-tuned chat models.
\newblock \emph{CoRR}, abs/2307.09288.

\bibitem[{Vaswani et~al.(2017)Vaswani, Shazeer, Parmar, Uszkoreit, Jones, Gomez, Kaiser, and Polosukhin}]{vaswani-nips-2017-attention}
Ashish Vaswani, Noam Shazeer, Niki Parmar, Jakob Uszkoreit, Llion Jones, Aidan~N Gomez, {\L}ukasz Kaiser, and Illia Polosukhin. 2017.
\newblock Attention is all you need.
\newblock \emph{Advances in neural information processing systems}, 30.

\bibitem[{Wang et~al.(2024)Wang, Ren, Li, Zhao, Liu, and Wen}]{wang-emnlp-2024-rear}
Yuhao Wang, Ruiyang Ren, Junyi Li, Xin Zhao, Jing Liu, and Ji{-}Rong Wen. 2024.
\newblock {REAR:} {A} relevance-aware retrieval-augmented framework for open-domain question answering.
\newblock In \emph{Proceedings of the 2024 Conference on Empirical Methods in Natural Language Processing, {EMNLP} 2024, Miami, FL, USA, November 12-16, 2024}, pages 5613--5626. Association for Computational Linguistics.

\bibitem[{Wang et~al.(2025)Wang, Ren, Wang, Zhao, Liu, Wu, and Wang}]{wang-2025-arxiv-unvieling}
Yuhao Wang, Ruiyang Ren, Yucheng Wang, Wayne~Xin Zhao, Jing Liu, Hua Wu, and Haifeng Wang. 2025.
\newblock Unveiling knowledge utilization mechanisms in llm-based retrieval-augmented generation.
\newblock \emph{CoRR}, abs/2505.11995.

\bibitem[{Wu et~al.(2024{\natexlab{a}})Wu, Zhao, and Zheng}]{wu-nips-2024-cream}
Tong Wu, Yanpeng Zhao, and Zilong Zheng. 2024{\natexlab{a}}.
\newblock An efficient recipe for long context extension via middle-focused positional encoding.
\newblock In \emph{The Thirty-eighth Annual Conference on Neural Information Processing Systems}.

\bibitem[{Wu et~al.(2024{\natexlab{b}})Wu, Luo, Li, Pan, Vu, and Haffari}]{wu-arxiv-2024-continual}
Tongtong Wu, Linhao Luo, Yuan{-}Fang Li, Shirui Pan, Thuy{-}Trang Vu, and Gholamreza Haffari. 2024{\natexlab{b}}.
\newblock Continual learning for large language models: {A} survey.
\newblock \emph{CoRR}, abs/2402.01364.

\bibitem[{Xiao et~al.(2023)Xiao, Tian, Chen, Han, and Lewis}]{xiao-arxiv-2023-streaming}
Guangxuan Xiao, Yuandong Tian, Beidi Chen, Song Han, and Mike Lewis. 2023.
\newblock Efficient streaming language models with attention sinks.
\newblock \emph{CoRR}, abs/2309.17453.

\bibitem[{Xiong et~al.(2024)Xiong, Liu, Molybog, Zhang, Bhargava, Hou, Martin, Rungta, Sankararaman, Oguz, Khabsa, Fang, Mehdad, Narang, Malik, Fan, Bhosale, Edunov, Lewis, Wang, and Ma}]{xiong-naacl-2024-effective}
Wenhan Xiong, Jingyu Liu, Igor Molybog, Hejia Zhang, Prajjwal Bhargava, Rui Hou, Louis Martin, Rashi Rungta, Karthik~Abinav Sankararaman, Barlas Oguz, Madian Khabsa, Han Fang, Yashar Mehdad, Sharan Narang, Kshitiz Malik, Angela Fan, Shruti Bhosale, Sergey Edunov, Mike Lewis, Sinong Wang, and Hao Ma. 2024.
\newblock Effective long-context scaling of foundation models.
\newblock In \emph{Proceedings of the 2024 Conference of the North American Chapter of the Association for Computational Linguistics: Human Language Technologies (Volume 1: Long Papers), {NAACL} 2024, Mexico City, Mexico, June 16-21, 2024}, pages 4643--4663. Association for Computational Linguistics.

\bibitem[{Xu et~al.(2024)Xu, Li, Tao, Shen, Cheng, Li, Xu, Tao, and Zhou}]{Xu-arxiv-2024-Survey}
Xiaohan Xu, Ming Li, Chongyang Tao, Tao Shen, Reynold Cheng, Jinyang Li, Can Xu, Dacheng Tao, and Tianyi Zhou. 2024.
\newblock A survey on knowledge distillation of large language models.
\newblock \emph{CoRR}, abs/2402.13116.

\bibitem[{Zeng et~al.(2024)Zeng, Xu, Wang, Zhang, Yin, Rojas, Feng, Zhao, Lai, Yu, Wang, Sun, Zhang, Cheng, Gui, Tang, Zhang, Li, Zhao, Wu, Zhong, Liu, Huang, Zhang, Zheng, Lu, Duan, Zhang, Cao, Yang, Tam, Zhao, Liu, Xia, Zhang, Gu, Lv, Liu, Liu, Yang, Song, Zhang, An, Xu, Niu, Yang, Li, Bai, Dong, Qi, Wang, Yang, Du, Hou, and Wang}]{Zeng-arxiv-2024-glm}
Aohan Zeng, Bin Xu, Bowen Wang, Chenhui Zhang, Da~Yin, Diego Rojas, Guanyu Feng, Hanlin Zhao, Hanyu Lai, Hao Yu, Hongning Wang, Jiadai Sun, Jiajie Zhang, Jiale Cheng, Jiayi Gui, Jie Tang, Jing Zhang, Juanzi Li, Lei Zhao, Lindong Wu, Lucen Zhong, Mingdao Liu, Minlie Huang, Peng Zhang, Qinkai Zheng, Rui Lu, Shuaiqi Duan, Shudan Zhang, Shulin Cao, Shuxun Yang, Weng~Lam Tam, Wenyi Zhao, Xiao Liu, Xiao Xia, Xiaohan Zhang, Xiaotao Gu, Xin Lv, Xinghan Liu, Xinyi Liu, Xinyue Yang, Xixuan Song, Xunkai Zhang, Yifan An, Yifan Xu, Yilin Niu, Yuantao Yang, Yueyan Li, Yushi Bai, Yuxiao Dong, Zehan Qi, Zhaoyu Wang, Zhen Yang, Zhengxiao Du, Zhenyu Hou, and Zihan Wang. 2024.
\newblock Chatglm: {A} family of large language models from {GLM-130B} to {GLM-4} all tools.
\newblock \emph{CoRR}, abs/2406.12793.

\bibitem[{Zhang et~al.(2024)Zhang, Zhang, Li, Zeng, Yang, Zhang, Wang, Tan, Li, and Liu}]{zhang-arxiv-2024-easycontext}
Peiyuan Zhang, Kaichen Zhang, Bo~Li, Guangtao Zeng, Jingkang Yang, Yuanhan Zhang, Ziyue Wang, Haoran Tan, Chunyuan Li, and Ziwei Liu. 2024.
\newblock Long context transfer from language to vision.
\newblock \emph{CoRR}, abs/2406.16852.

\bibitem[{Zhao et~al.(2023)Zhao, Zhou, Li, Tang, Wang, Hou, Min, Zhang, Zhang, Dong, Du, Yang, Chen, Chen, Jiang, Ren, Li, Tang, Liu, Liu, Nie, and Wen}]{zhao-arxiv-2023-survey}
Wayne~Xin Zhao, Kun Zhou, Junyi Li, Tianyi Tang, Xiaolei Wang, Yupeng Hou, Yingqian Min, Beichen Zhang, Junjie Zhang, Zican Dong, Yifan Du, Chen Yang, Yushuo Chen, Zhipeng Chen, Jinhao Jiang, Ruiyang Ren, Yifan Li, Xinyu Tang, Zikang Liu, Peiyu Liu, Jian{-}Yun Nie, and Ji{-}Rong Wen. 2023.
\newblock A survey of large language models.
\newblock \emph{CoRR}, abs/2303.18223.

\bibitem[{Zhu et~al.(2024)Zhu, Yang, Wang, Song, Wu, Wei, and Li}]{zhu-iclr-2024-pose}
Dawei Zhu, Nan Yang, Liang Wang, Yifan Song, Wenhao Wu, Furu Wei, and Sujian Li. 2024.
\newblock Pose: Efficient context window extension of llms via positional skip-wise training.
\newblock In \emph{The Twelfth International Conference on Learning Representations, {ICLR} 2024, Vienna, Austria, May 7-11, 2024}. OpenReview.net.

\end{thebibliography}
\newpage
\appendix

\section{Skipped Positional Indices}
\label{app:cream}

Skipped positional indices are widely employed for effective long-context training. The method simulates long-distance dependency via modifying the positional indices. In this section, we introduce two skipped positional indices employed in our methods.

\subsection{CREAM}
Continuity-Relativity indExing with gAussian Middle (CREAM)~\cite{wu-nips-2024-cream} is a method that modifies position indices to simulate long positions. The main idea is to make the model better focused on the middle part of the long sequences. 

As described in Section~\ref{sec:s2l_distillation}, the positional indices are first split into three non-overlapped parts, \ie $\bm p_{head}$, $\bm p_{tail}$, and $\bm p_{tail}$, where the length of the head and the tail are $T_b$. There are two segmentation methods and we randomly select one method to determine the number of $T_b$. To achieve the continuity of the middle part, $T_b$ is set as a small number, where we set it to four times the context window extension factor. To learn more relative positions, $T_b$ is set as one-third of the input length $T/3$. 

To better concentrate on the middle positions, the method proposes a truncated Gaussian function to determine the end position of the middle part $p_e$. Initially, the method defines a Gaussian distribution $f(x)$ and its cumulative distribution function $F(x)$:
\begin{align}
    f(x) &= \frac 1 {\sigma \sqrt 2\pi} \exp(-\frac {(x-\mu)^2}{2\sigma^2}), \\
    F(x) &= \int_{-\infty}^x f(t) dt,
\end{align}
where $\mu= 1+T_l/T$ and $\sigma=3$. Then, the method samples 1000 points $x_i$ uniformly from 1 to extension factor $T_l/T$ and calculates their cumulative distribution function $F(x_i)$. After that, the method samples $F(u)$ from the Gaussian distribution and uses interpolation and inverse transformation to obtain the corresponding $u$. The number $u$ then rounds to the nearest integer $\alpha$.
\begin{align}
    u &= x_{i-1} + \frac{(x_i-x_{i-1})(F(u)-F(x_{i-1}))}{F(x_i)-F(x_{i-1})},\\
    \alpha &= \operatorname{round}(u),
\end{align}
where $F(x_i)$ and $F(x_{i-1})$ are the two nearest numbers to $F(u)$ and $F(u)$ is sampled uniformly from 0 to 1. Finally, we sample an integer as the end of the middle part:
\begin{align}
    T_{me} &\sim \operatorname{Uniform}(T_b+\alpha (T_l-2T_b), \alpha T-T_b-1),\\
    T_{mb} &= T_{me}+2T_b-T.
\end{align}

\subsection{Uniform Sampling}

In the uniform sampling method, we keep the same as the CREAM method except the sampling method of the positions of the middle part. Instead, we employ a uniform sampling method to determine the end positions of the middle part, as formulated as follows: 
\begin{align}
    T_{me} &\sim \operatorname{Uniform}(T-T_b, T_l-T_b-1),\\
    T_{mb} &= T_{me}+2T_b-T,
\end{align}
where $\operatorname{Uniform}$ denotes uniformly sampling from this range.
\section{Training Details}
\label{app:train_details}

\subsection{Training Datasets}

In our proposed method, we employ three datasets: $\mathcal{D}_1$, $\mathcal{D}_2$, and $\mathcal{D}_3$, which are specifically designed for long-text training, short-text distillation, and short-to-long distillation, respectively. Following the sampling strategy of~\citet{fu-icml-2024-data}, we sample data in the same proportion as Llama-2 from the SlimPajama datasets~\cite{soboleva-huggingface-2023-slimpajama}, tokenize the samples, and pack them to the target length. To enhance the preservation of short-text capabilities during the short-text distillation phase, we adopt the approach proposed by~\cite{gao-arxiv-2024-how}, utilizing higher-quality datasets. Specifically, we select Fineweb-Edu~\cite{penedo-arxiv-2024-fineweb}, Stack-v2~\cite{Lozhkov-arxiv-2024-stack}, the mathematical subset of Fineweb-Edu~\cite{penedo-arxiv-2024-fineweb}, Proof-Pile-2~\cite{Azerbayev-ICLR-2024-proof}, PG19~\cite{rae-iclr-2020-pg19}, Arxiv~\cite{soboleva-huggingface-2023-slimpajama}, and Wikipedia~\cite{soboleva-huggingface-2023-slimpajama}. These datasets are carefully chosen to balance quality and diversity.
Additionally, we incorporate instruction-tuned datasets that have been demonstrated to be beneficial for model performance~\cite{hu-arxiv-2024-minicpm}, including Tulu-v2-sft-mixture\footnote{\url{https://huggingface.co/datasets/allenai/tulu-v2-sft-mixture}}, MathInstruct\footnote{\url{https://huggingface.co/datasets/TIGER-Lab/MathInstruct}}, WizardLM-evol-instruct-V2\footnote{\url{https://huggingface.co/datasets/WizardLMTeam/WizardLM\_evol\_instruct\_V2\_196k}}, and Magicoder-Evol-Instruct-110K\footnote{\url{https://huggingface.co/datasets/ise-uiuc/Magicoder-Evol-Instruct-110K}}. The detailed proportions of these datasets utilized in our method are summarized in Table~\ref{tab:data}.

\begin{table}[htb]
    \centering
    \resizebox{\linewidth}{!}{
    \begin{tabular}{cc}
    \toprule
         Dataset& Data Mixture\\\midrule
         $\mathcal D_1$& \tabincell{c}{Wikipedia: 0.034, CommonCrawl: 0.534, \\StackExchange: 0.032, C4: 0.266, \\Github: 0.050, ArXiv: 0.043, Book: 0.041}\\\midrule
         $\mathcal D_3$& \tabincell{c}{Wikipedia: 0.034, CommonCrawl: 0.534, \\StackExchange: 0.032, C4: 0.266, \\Github: 0.050, ArXiv: 0.043, Book: 0.041}\\\midrule
         $\mathcal D_2$& \tabincell{c}{Fineweb-Edu: 0.476, Fineweb-Edu-Math: 0.722\\ Proof-Pile-2: 0.230, Stack-v2: 0.095, PG19: 0.095, \\Wikipedia: 0.096, ArXiv: 0.095, Instructions: 0.047}\\\bottomrule
    \end{tabular}}
    \caption{Data Mixture of three datasets.}
    \label{tab:data}
\end{table}

\subsection{Evaluated Models}
We select Llama-3-8B and Mistral-7B-v0.3 as our evaluated models. For Llama-3-8B, we extend its context window to 32K and 128K, with the ABF and PI methods. For Mistral-7B-v0.3, we extend its context window to 128K with ABF method. The configuration and training length of these models are displayed in Table~\ref{tab:model_config}.

\begin{table}[htb]
    \centering
    \resizebox{\linewidth}{!}{
    \begin{tabular}{lcccccc}
    \toprule
         Model&  CW&  PE&  Ratio&  $L_{\mathcal D_1}$&  $L_{\mathcal D_2}$&  $L_{\mathcal D_3}$\\\midrule
         Llama-3-8B&  32K&  ABF&  2e7&  32000&  1024&  8192\\
         Llama-3-8B&  32K&  PI&  4&  32000&  1024&  8192\\
         Llama-3-8B&  128K&  ABF&  1e8&  128000&  1024&  8192\\
         Mistral-7B-v0.3&  128K&  ABF&  2e7&  128000&  1024&  32768\\
         \bottomrule
    \end{tabular}}
    \caption{Configuration of models and length of datasets. CW denotes the context window of target models, PE denotes the scaling methods of RoPE, Ratio denotes the RoPE theta for ABF methods, and the interpolation ratio for PI methods. $L_{\mathcal D_1}$, $L_{\mathcal D_2}$, and $L_{\mathcal D_3}$ denotes the length of the three datasets.}
    \label{tab:model_config}
\end{table}

\subsection{Training Configurations}
We employ ring flash attention~\cite{liu-arxiv-2023-ringattn} in EasyContext framework~\cite{zhang-arxiv-2024-easycontext} (Apache-2.0 License) to train our models with 8 A800 GPUs. The learning rate is  fixed as 2e-5 without warmup and the training tokens of each batch are about 2M tokens. We employ AdamW optimizer~\cite{Loshchilov-iclr-2019-adamw} with the weight decay of 0.1, $\beta_1$ of 0.9, and $\beta_2$ of 0.95.  The hyper-parameters $\alpha_1$ and $\alpha_2$ are set as $\alpha_1=5, \alpha_2=10$ for 32K, and $\alpha_1=2, \alpha_2=15$ for 128K. All the models are trained with 512 steps with about 1B tokens.

\subsection{Training Costs}

We report the training costs of our method and other baselines. All the models are trained on 8 A800 GPUs with ring attention. The time costs for training these models are shown in Table~\ref{tab:time}. Compared with baselines, our methods will cost about $10\%$ extra computation than the only continually pre-training on the same data. Additionally, comparing only training on long texts, LongReD costs similar time in 32K while saving about $20\%$ computations in 128K settings.
\begin{table}[htb]
    \centering
    \resizebox{\linewidth}{!}{
    \begin{tabular}{l|cccc} \toprule
         Model&  CW&  Data&  Method&  Time(h)\\ \midrule
 \multirow{6}{*}{Llama-3-8B}& 32K& Long& CPT& 22.4\\ 
 & 32K& Mix& CPT&19.9\\ 
 & 32K& Mix& LongReD& 22.4\\
 & 128K& Long& CPT& 42.8\\
 & 128K& Mix& CPT& 30.4\\
 & 128K& Mix& LongReD& 33.4\\\bottomrule
    \end{tabular}}
    \caption{Training time of Llama-3-8B with different methods and target context windows.}
    \label{tab:time}
\end{table}

\section{Evaluation Details}
\label{app:evaluation}

To evaluate the performance of LLMs on short-text and long-text tasks, we employ multiple benchmarks. In this section, we introduce the datasets for evaluating different capacities:
\begin{itemize}
    \item \emph{General}: MMLU~\cite{Hendrycks-iclr-2021-mmlu}, BBH~\cite{suzgun-acl-2023-bbh}, LAMBADA~\cite{paperno-acl-2016-lambada}
    \item \emph{Math}: MATH~\cite{henderycksnips-2021-math}, GSM8K~\cite{cobbe-arxiv-2021-gsm8k}
    \item \emph{Coding}: HumanEval~\cite{chen-arxiv-2021-humaneval}, MBPP~\cite{austin-arxiv-2021-mbpp}
    \item \emph{Reading comprehension}: Squadv2~\cite{rajpurkar-EMNLP-2016-squad}, Quac~\cite{choi-acl-2018-quac}, TriviaQA~\cite{Joshi-acl-2017-triviaqa}, Drop~\cite{dua-naacl-2019-drop}, BoolQ~\cite{clark-naccl-2019-boolq}
    \item \emph{Commonsense question answering}: Openbookqa~\cite{todor-emnlp-2018-openbookqa}, Commonsenseqa~\cite{talmor-naacl-2019-commonsenseqa}, ARC-C~\cite{clark-arxiv-2018-arc}, SIQA~\cite{Sap-emnlp-2019-siqa}, PIQA~\cite{bisk-aaai-2020-piqa}),
    \item \emph{Long text processing}: RULER~\cite{Hsieh-arxiv-2024-RULER}
\end{itemize}
we also present details of our evaluation configurations, as shown in Table~\ref{tab:evaluation_details}.

\begin{table}[htb]
    \centering
    \small{
    \begin{tabular}{llcc}
    \toprule
         Dataset&   CoT&\#Shots& Metric\\\midrule
         MMLU&   $\times$&5& Probability\\
         BBH&   \checkmark&3& EM\\
         Lambada&   $\times$&0& accuracy\\
         MBPP&   $\times$&3& pass@1\\
         HumanEval&  $\times$ &0& pass@1\\
         GSM8K&  \checkmark &4& accuracy\\
         MATH&  \checkmark &4& accuracy\\
         PIQA&   $\times$&0& Probability\\
 SIQA&  $\times$&0&Probability\\
 OpenBookQA&  $\times$&0&Probability\\
 ARC(C)&  $\times$&25&Probability\\
 CommonSenseQA&  $\times$&7&PPL\\
 Quac& $\times$& 1&F1\\
 TriviaQA& $\times$& 5&F1\\
 DROP&  $\times$&3&F1\\
 BoolQ&  $\times$&3&F1\\
 Squad-v2&  $\times$&1&F1\\
 RULER&  $\times$&0&Accuracy\\
    \bottomrule
    \end{tabular}}
    \caption{Configurations of evaluated benchmarks. CoT denotes employing the Chain-of-Thought prompt, \#Shot denotes the number of shots in prompts, and Metric denotes the evaluation metric for the benchmark.}
    \label{tab:evaluation_details}
\end{table}
\begin{table*}[htb]
    \centering
    \resizebox{\linewidth}{!}{
    \begin{tabular}{lllllccc}
    \toprule
         Model&  RoPE&  CW&  Base Model&  Tokens&  MMLU&  MMLU Ratio& Simi\\\midrule
 \multicolumn{8}{c}{GradientAI Models}\\\midrule
         Llama-3-8B-Ins-G-262k&  ABF(2.8e8)&  262K&  -&  -&  61.9&  0.945& 0.920\\
         Llama-3-8B-Ins-G-1048k&  ABF(3.6e9)&  1048K&  -&  -&  59.7&  0.911& 0.900\\
         Llama-3-8B-Ins-G-4194k&  ABF(4.5e10)&  4194K&  -&  -&  56.1&  0.856& 0.872\\\midrule
         \multicolumn{8}{c}{Our Models (Directly Extension)}\\\midrule
 Llama-3-32K-5e6& ABF(5e6)& 32K& Llama-3-8B& 256M& 62.2& 0.960&0.943\\
 Llama-3-32K-5e7& ABF(5e7)& 32K& Llama-3-8B& 256M& 62.0& 0.957&0.934\\
 Llama-3-32K-5e8& ABF(5e8)& 32K& Llama-3-8B& 256M& 60.9& 0.940&0.924\\
         Llama-3-32K-5e9&  ABF(5e9)&  32K&  Llama-3-8B& 256M&  60.4&  0.932& 0.915\\
         Llama-3-32K-5e10&  ABF(5e10)&  32K&  Llama-3-8B& 256M&  59.8&  0.923& 0.908\\
 Llama-3-32K-5e20& ABF(5e20)& 32K& Llama-3-8B& 256M& 56.5& 0.872&0.856\\
 Llama-3-32K-PI4& PI(4)& 32K& Llama-3-8B& 256M& 60.9& 0.940&0.917\\
         Llama-3-32K-PI16&  PI(16)&  32K&  Llama-3-8B& 256M&  56.1&  0.866& 0.862\\\midrule
 \multicolumn{8}{c}{Our Models (Gradually Extension)}\\\midrule
 Llama-3-16K-5e6& ABF(5e6)& 16K& Llama-3-8B& 256M& 62.3& 0.961&0.946\\
 Llama-3-32K-5e7& ABF(5e7)& 32K& Llama-3-16K-5e6& 256M& 60.7& 0.938&0.927\\
 Llama-3-64K-5e8& ABF(5e8)& 64K& Llama-3-32K-5e7& 256M& 60.3& 0.930&0.912\\
 Llama-3-128K-5e9& ABF(5e9)& 128K& Llama-3-64K-5e8& 256M& 57.2& 0.883&0.901\\\bottomrule
    \end{tabular}}
    \caption{Details of evaluated models for the analysis of the relationship between distribution drift and short-text performances. }
    \label{tab:drift}
\end{table*}
The details of metrics employed to evaluate the performance are described as follows:
\begin{itemize}
    \item \emph{Probability}: For a choice task, all the choices are provided in the prompt and the Probability of each choice is calculated. The choice with the maximum Probability will be compared with the correct choice and the accuracy is reported.
    \item \emph{PPL}: For a choice task, the choices are not provided. Instead, the content of each choice is input after the question, and the average perplexity of each choice is calculated. The choice with the lowest PPL is selected as the correct answer.
    \item \emph{Pass@1}:  For coding tasks, the model is asked to give one solution. Then, we evaluate the solution with the unit testing and count the numbers that pass the testing. 
    \item \emph{EM}:  For the question-answering task, the exact matching metric measures whether the output is the same as the golden answer.
    \item \emph{F1}:  For the question-answering task, the F1 first splits the golden answer and output of the model into words, and calculates the F1 score of these word lists. 
    \item \emph{Accuracy}:   Accuracy measures the number of correctly answered samples out of all the samples. For RULER, the sample is correct if the answer is in the output or each part of the answer (only for qa1 and qa2).  

\end{itemize}

We evaluate these models using the LLMBox~\cite{tang-arxiv-2024-llmbox} library for most benchmarks, and OpenCompass~\cite{2023opencompass} for the math-related benchmarks.

\section{Models for Evaluating Distribution Drift}
\label{app:drift_models}
To evaluate the relationship between the distribution drift and the short performance, we first employ open-sourced three long-context models from Gradient AI~\cite{gradient-hf-2024-longcontextllama3}, which are continually pre-trained from Llama-3-8B-Instruct with only several hundred million tokens. Then, we train Llama-3-8B with ABF and PI methods with different extension ratios on 256M tokens with a length of 32K. We also follow~\citet{gradient-hf-2024-longcontextllama3}, gradually extending the context window length to twice its original length. For each extension, we train the model with 256M tokens and 
enlarge the RoPE theta ten times. The full model list and their MMLU scores and hidden states similarity can be seen in Table~\ref{tab:drift}. Typically, a larger extension ratio will cause a greater shift in the model's distribution, further leading to a decline in the model's performance on short texts. We provide theoretical proof of the impact of the RoPE bases on the drifts of attention distributions.

\subsection{Proof of RoPE Base Impact on Attention Scores}
Given a query $\mathbf{q}^m$ and a key $\mathbf{k}^n$, their inner product with RoPE encoding can be expressed as:

\begin{align}
   & (\mathbf{R}^d_{\Theta, m} \mathbf{q}^m)^\top (\mathbf{R}^d_{\Theta, n} \mathbf{k}^n)\\ =& \mathrm{Re}\left[\sum_{i=0}^{d/2-1} \mathbf{q}^m_{[2i:2i+1]} {\mathbf{k}^n_{[2i:2i+1]}}^* e^{\mathrm{i}(m-n)\theta_i}\right]
\end{align}

where $[2i:2i+1]$ denotes the $i$-th subspace. Following the definition from \citet{su-neurocomputing-2024-roformer}, we let $h_i = \mathbf{q}^m_{[2i:2i+1]} {\mathbf{k}^n_{[2i:2i+1]}}^*$ and $S^\Theta_j = \sum_{k=0}^{j-1} e^{\mathrm{i}(m-n)\theta_k}$. We extend these definitions with $h_{d/2} = 0$ and $S^\Theta_0 = 0$. Applying Abel transformation, we rewrite the summation:

\begin{align}
\sum_{i=0}^{d/2-1} h_i e^{\mathrm{i}(m-n)\theta_i} 
&= \sum_{i=0}^{d/2-1} h_i (S^\Theta_{i+1} - S^\Theta_i) \\
&= -\sum_{i=0}^{d/2-1} S^\Theta_{i+1}(h_{i+1} - h_i)
\end{align}

\textbf{Theorem.} Let $\Theta = b^{-2i/d}$ and $\Theta' = {b'}^{-2i/d}$ be two RoPE bases with $b > b' > 1$. For fixed $\mathbf{q}^m$ and $\mathbf{k}^n$, their attention score difference satisfies:
\[
\left| \textit{Attn}(b) - \textit{Attn}(b') \right| \leq C \cdot \sum_{i=0}^{d/2-1} \left(|S^{\Theta}_{i+1}| + |S^{\Theta'}_{i+1}|\right)
\]
where $C = \max\limits_{0 \leq i < d/2} |h_{i+1} - h_i|$.

\textbf{Proof.}  
The attention scores under different bases are:
\begin{align}
    \textit{Attn}(b) = \mathrm{Re}\left[-\sum_{i=0}^{d/2-1} S^\Theta_{i+1}(h_{i+1} - h_i)\right],\\ \textit{Attn}(b') = \mathrm{Re}\left[-\sum_{i=0}^{d/2-1} S^{\Theta'}_{i+1}(h_{i+1} - h_i)\right]
\end{align}

Their difference satisfies:
\begin{align}
&\left| \textit{Attn}(b) - \textit{Attn}(b') \right|\\ 
\leq & \left|\sum_{i=0}^{d/2-1} \left(S^\Theta_{i+1} - S^{\Theta'}_{i+1}\right)(h_{i+1} - h_i)\right| \\
\leq & \sum_{i=0}^{d/2-1} \left| S^\Theta_{i+1} - S^{\Theta'}_{i+1}\right| \cdot \left| h_{i+1} - h_i\right| \\
\leq & C \sum_{i=0}^{d/2-1} \left(|S^{\Theta}_{i+1}| + |S^{\Theta'}_{i+1}|\right)
\end{align}
where the last inequality follows from $|S^\Theta_{i+1} - S^{\Theta'}_{i+1}| \leq |S^\Theta_{i+1}| + |S^{\Theta'}_{i+1}|$ by triangle inequality.

To analyze the expected difference across positions, consider:
\begin{align}
&\mathbb{E}\left[\left| \textit{Attn}(b) - \textit{Attn}(b') \right|\right] \\
\leq & \frac{1}{N} \sum_{q,k,t} C^{q,k} \sum_{i=0}^{d/2-1} \left(|S^{\Theta,t}_{i+1}| + |S^{\Theta',t}_{i+1}|\right) \\
\leq & \frac{C}{N} \sum_{t=0}^{T} (T-t) \sum_{i=0}^{d/2-1} \left(|S^{\Theta,t}_{i+1}| + |S^{\Theta',t}_{i+1}|\right)
\end{align}

\begin{table*}[htb]
    \centering
    \begin{tabular}{cccccccccc}
    \toprule
         Base&  5e5&  5e6&  5e7&  5e8&  5e9&  5e10&  5e11&  5e12& 5e13\\\midrule
         $B(\Theta)$&  1.3e6&  1.5e6&  1.9e6&  2.3e6&  3.0e6&  3.3e6&  3.6e6&  3.9e6& 4.1e6\\\bottomrule
    \end{tabular}
    \caption{Upper bound of different RoPE bases.}
    \label{tab:upbound}
\end{table*}

where $T$ is the maximum sequence length and $t = |m-n|$ is the relative position. Let $B(\Theta) = \sum_{t=0}^{T} (T-t) \sum_{i=0}^{d/2-1} |S^{\Theta,t}_{i+1}|$. Previous work has demonstrated that RoPE has long-term decay and a large base will lead to slow attention decay. For bases $b_3 > b_2 > b_1$, we have $B(\Theta_3) > B(\Theta_2) > B(\Theta_1)$ due to slower attention decay with larger bases. As shown in Table~\ref{tab:upbound}, empirical measurements on Llama-3-8B confirm that $B(\Theta)$ increases with $b$, proving that larger bases induce greater attention score changes during context window extension.

\section{Analysis of Distillation Length}
\label{app:latent}

\begin{table*}
    \centering
    \resizebox{\textwidth}{!}{
    \begin{tabular}{l|ccccccccccc} \toprule
         Model&  CW&  PE&  Data&  Method&  MMLU&  BBH&  LAMBADA&  HUMANEVAL&   MBPP&GSM8K&MATH\\\midrule
 \multirow{13}{*}{Llama-3-8B}& 8K& -& -& -& 64.80& 63.97& 75.78& 34.75& 48.28& 50.04&10.20\\ \cmidrule{2-12}
 & 32K& ABF& Long& CPT& 62.00& 59.90& 74.27& 14.02& 44.62& 49.20&9.80\\ 
 & 32K& ABF& Mix& CPT&62.60& 60.54& 74.97& 16.46& 38.34& 49.43&9.20\\
 & 32K& ABF& Mix& LongReD-C& 64.40& 62.78& 75.35& 35.37& 44.42& 52.84&10.20\\
 & 32K& ABF& Mix& LongReD-U& 64.40& 61.75& 75.02& 34.76& 44.66& 52.69&9.20\\\cmidrule{2-12}
 & 32K& PI& Long& CPT& 61.40& 56.20& 73.92& 12.20& 42.06& 44.96&7.40\\ 
 & 32K& PI& Mix& CPT& 60.50& 57.17& 74.99& 16.46& 38.12& 49.13&9.20\\ 
 & 32K& PI& Mix& LongReD-C& 63.30& 60.83& 75.22& 34.76& 42.02& 51.55&10.00\\
 \cmidrule{2-12}
 & 128K& ABF& Long& CPT& 61.40& 56.22& 74.56& 17.07& 43.00& 47.23&8.20\\
 & 128K& ABF& Mix& CPT& 62.40& 61.62& 75.16& 21.95& 6.12& 50.04&8.00\\
 & 128K& ABF& Mix& LongReD-C& 63.60& 60.25& 74.11& 35.37& 44.68& 52.92&7.60\\
  & 128K& ABF& Mix& LongReD-U& 63.60& 59.45& 73.78& 35.96& 43.78& 49.13&10.20\\\midrule
 \multirow{5}{*}{\makecell{Mistral\\-7B-v0.3}}& 32K& -& -& -& 62.30& 58.49& 75.28& 24.39& 42.00& 45.56& 8.40\\\cmidrule{2-12}
  & 128K& ABF& Long& CPT& 51.90&	43.3&	70.68 & 21.34	&31.4&26.38	&5.20\\
 & 128K& ABF& Mix& CPT& 54.70	&44.87	&68.39 & 19.51&	33.2 & 27.60 &	3.00\\
 & 128K& ABF& Mix& LongReD-C& 58.90&	51.58	&73.86 & 20.12	&33.2 & 33.36 &6.40\\
 & 128K& ABF& Mix& LongReD-U& 60.1&	52.08&	75.1& 23.78	&35.5& 34.12&	5.00\\
 \bottomrule
    \end{tabular}}
    \caption{Comparison of performances of short-text and long-text benchmarks of our methods with other baselines. CW denotes the context window length, PE denotes the scaling method of RoPE, RC denotes reading comprehension, and Common denotes commonsense question answering. }
    \label{tab:results_detail1}
\end{table*}

\begin{figure}[htb]
    \centering
    \includegraphics[width=\linewidth]{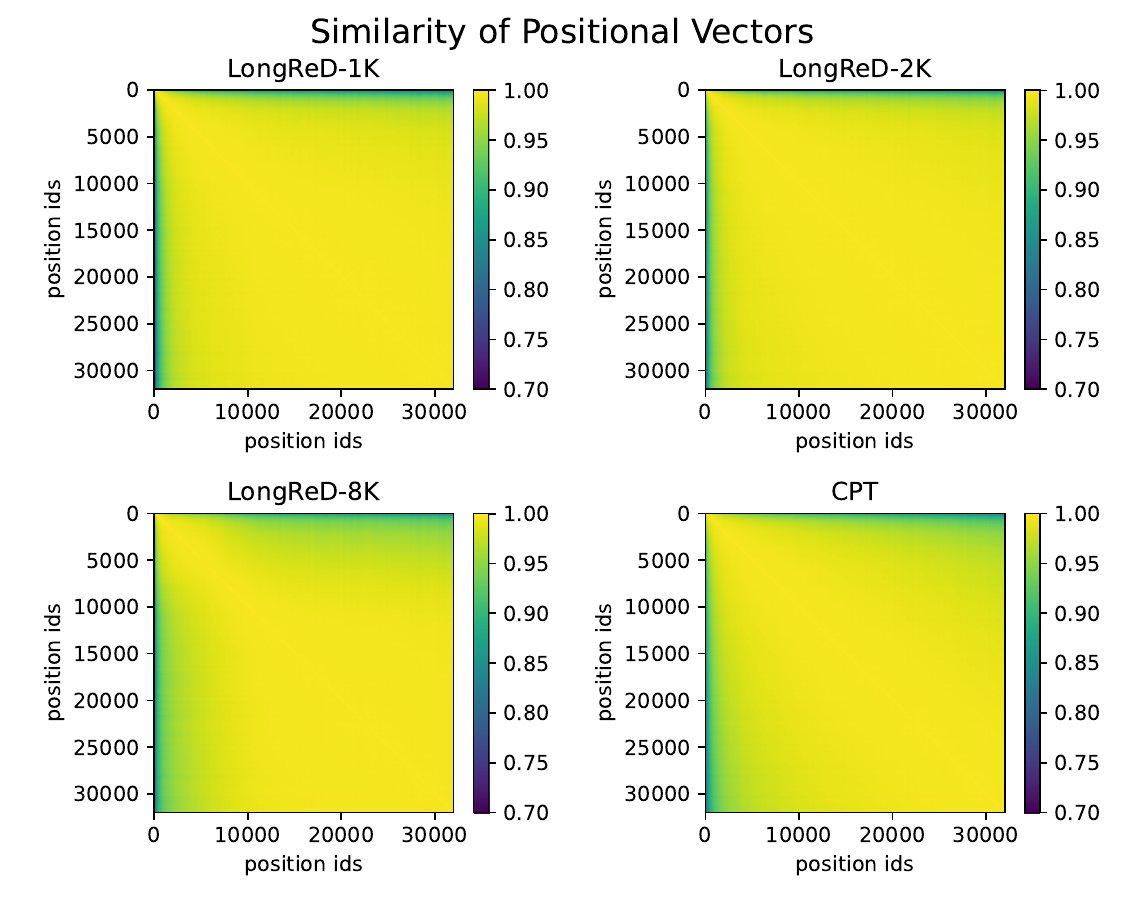}
    \caption{Simialrity matrices of positional vectors inside and outside the context window.}
    \label{fig:pe}
\end{figure}

As discussed in Section~\ref{sec:ablation}, increasing the distillation length will harm the long text modeling capacities. To explore the reason for the performance decline, we employ positional vectors~\cite{dong-arxiv-2024-exploring}, a method to extract latent positional information from the hidden states. Specifically, given the hidden states from different samples, we average the hidden states at the same position to obtain vectors that 
representing positional information, denoted as positional vectors:
\begin{equation}
   \bm  p_{l,i} = \frac 1 N\sum_{s=1}^N \bm h_{l,i}^s 
\end{equation}
Then, we compute the cosine similarities of positional vectors with different places. All the positional vectors are calculated on samples from SlimPajama~\cite{soboleva-huggingface-2023-slimpajama} with a length of 32000 tokens. The similarity matrices of the positional vectors of models with distillation lengths of 1024, 2048, and 8192 as well as the models trained with only long texts are shown in Figure~\ref{fig:pe}. We can observe that a large distillation length will cause a significant discontinuity in the implicit positional information inside and outside the original context window.

\section{Results Details}
\label{app:all_results}
In this section, we display the details of performances on these benchmarks. The results of benchmarks evaluating general, coding, and math capacities are shown in Table~\ref{tab:results_detail1}. Table~\ref{tab:results_detail2} and Table~\ref{tab:results_detail3} show the performance on commonsense question answering and reading comprehension benchmarks respectively.

\begin{table*}
    \centering
    \resizebox{\textwidth}{!}{
    \begin{tabular}{l|ccccccccc} \toprule
         Model&  CW&  PE&  Data&  Method&  CommonsenseQA&  OpenBookQA&  PIQA&  SIQA&   ARC-C\\\midrule
 \multirow{13}{*}{Llama-3-8B}& 8K& -& -& -& 73.22& 71& 77.97& 61& 79.52\\ \cmidrule{2-10}
 & 32K& ABF& Long& CPT& 71.50& 66.40& 74.1& 59.42& 77.73\\ 
 & 32K& ABF& Mix& CPT&72.65& 69.60& 77.09& 62.74& 77.30\\
 & 32K& ABF& Mix& LongReD-C& 72.56& 71.80& 77.80& 60.34& 79.27\\
 & 32K& ABF& Mix& LongReD-U& 72.89& 71.80& 77.91& 61.46& 79.01\\\cmidrule{2-10}
 & 32K& PI& Long& CPT& 71.25& 62.80& 72.03& 55.58& 76.88\\ 
 & 32K& PI& Mix& CPT& 70.19& 69.00& 77.04& 62.28& 76.96\\ 
 & 32K& PI& Mix& LongReD-C& 71.91& 70.80& 75.90& 59.98& 77.90\\
 \cmidrule{2-10}
 & 128K& ABF& Long& CPT& 65.19& 64.00& 77.15& 59.83& 75.68\\
 & 128K& ABF& Mix& CPT& 72.65& 70.00& 77.53& 61.62& 77.73\\
  & 128K& ABF& Mix& LongReD-C& 72.40& 69.20& 76.01& 61.16& 78.75\\
 & 128K& ABF& Mix& LongReD-U& 72.40& 69.80& 77.04& 61.00& 77.99\\\midrule
 \multirow{5}{*}{\makecell{Mistral\\-7B-v0.3}}& 32K& -& -& -& 71.50	&66.8	&66.65	&56.4	&79.01\\\cmidrule{2-10}
  & 128K& ABF& Long& CPT& 66.99&	34.8	&56.8&	38.89	&67.15\\
 & 128K& ABF& Mix& CPT& 66.99&	34.8	&56.8	&38.89&	67.15\\
 & 128K& ABF& Mix& LongReD-C& 69.04	&65.4	&62.68	&53.89&	75.26\\
& 128K& ABF& Mix& LongReD-C& 70.84	&65.6	&63.87	&55.48	&78.5\\
 \bottomrule
    \end{tabular}}
    \caption{Comparison of performances of short-text and long-text benchmarks of our methods with other baselines. CW denotes the context window length, PE denotes the scaling method of RoPE, RC denotes reading comprehension, and Common denotes commonsense question answering. }
    \label{tab:results_detail2}
\end{table*}

\begin{table*}
\centering
    \resizebox{\textwidth}{!}{
    \begin{tabular}{l|ccccccccc} \toprule
         Model&  CW&  PE&  Data&  Method&  SquadV2&  quacQ&  TriviaQA&  BoolQ&   DROP\\\midrule
 \multirow{13}{*}{Llama-3-8B}& 8K& -& -& -& 72.06& 35.65& 75.42& 82.6& 53.18\\ \cmidrule{2-10}
 & 32K& ABF& Long& CPT& 66.56& 34.03& 70.67& 83.12& 50.52\\ 
 & 32K& ABF& Mix& CPT&0.89& 16.97& 72.60& 82.60& 0.17\\
 & 32K& ABF& Mix& LongReD-C& 74.18& 34.39& 74.24& 82.29& 49.10\\
 & 32K& ABF& Mix& LongReD-U& 72.45& 34.50& 74.16& 82.26& 49.18\\\cmidrule{2-10}
 & 32K& PI& Long& CPT& 73.26& 33.09& 70.95& 81.16& 51.29\\ 
 & 32K& PI& Mix& CPT& 7.53& 21.43& 72.55& 81.44& 0.21\\ 
 & 32K& PI& Mix& LongReD-C& 71.96& 34.73& 74.87& 82.94& 49.47\\
 \cmidrule{2-10}
 & 128K& ABF& Long& CPT& 66.01& 32.81& 73.13& 81.16& 49.35\\
 & 128K& ABF& Mix& CPT& 1.08& 18.41& 74.06& 82.66& 0.21\\
 & 128K& ABF& Mix& LongReD-C& 70.67& 33.97& 75.19& 82.51& 49.63\\
 & 128K& ABF& Mix& LongReD-U& 70.26& 34.85& 75.13& 82.84& 49.29\\\midrule
 \multirow{5}{*}{\makecell{Mistral\\-7B-v0.3}}& 32K& -& -& -& 72.47&	31.84	&76.1	&83.43&	45.46\\\cmidrule{2-10}
  & 128K& ABF& Long& CPT& 55.15	&29.07	&62.9	&80.83	&37.4\\
 & 128K& ABF& Mix& CPT& 53.92	&31.88	&63.85	&78.32&	35.73\\
 & 128K& ABF& Mix& LongReD-C& 70.45	&30.16	&68.16	&80.61&	41.9\\
 & 128K& ABF& Mix& LongReD-U& 72.6	&30.82&	71.84	&81.16&	43.5\\
 \bottomrule
    \end{tabular}}
    \caption{Comparison of performances of short-text and long-text benchmarks of our methods with other baselines. CW denotes the context window length, PE denotes the scaling method of RoPE, RC denotes reading comprehension, and Common denotes commonsense question answering. }
    \label{tab:results_detail3}
\end{table*}

\end{document}